\documentclass[11pt]{article}

\usepackage{jmlr2e}
\usepackage{url}
\pdfoutput=1
\usepackage{hyperref}
\hypersetup{
    colorlinks=true,
    linkcolor=black,
    citecolor=black,
    filecolor=black,
    urlcolor=black,
}

\usepackage{natbib}

\usepackage{tikz}
\usepackage{pgfplots, pgfplotstable}
\usetikzlibrary{pgfplots.groupplots}
\usetikzlibrary{plotmarks}
\pgfplotsset{compat=newest}
\usetikzlibrary{bayesnet}

\usepackage{tikzscale}
\usepackage{subcaption}
\usepackage{float}

\usepackage{bm}
\usepackage{amsmath}
\usepackage{amssymb}
\usepackage{graphicx}

\makeatletter
\def\input@path{{./}{../diagrams/}{../../../tex_inputs/}}
\makeatother
\graphicspath{{./},{../diagrams/}}
\usepackage{color}
\usepackage{verbatim}

\definecolor{brown}{rgb}{0.9,0.59,0.078}
\definecolor{ironsulf}{rgb}{0,0.7,.5}
\definecolor{lightpurple}{rgb}{0.156,0,0.245}

\ifdefined\blackBackground
\definecolor{colorOne}{rgb}{0, 1, 1}
\definecolor{colorTwo}{rgb}{1, 0, 1}
\definecolor{colorThree}{rgb}{1, 1, 0}
\definecolor{colorTwoThree}{rgb}{1, 0, 0}
\definecolor{colorOneThree}{rgb}{0, 1, 0}
\definecolor{colorOneTwo}{rgb}{0, 0, 1}
\else
\definecolor{colorOne}{rgb}{1, 0, 0}
\definecolor{colorTwo}{rgb}{0, 1, 0}
\definecolor{colorThree}{rgb}{0, 0, 1}
\definecolor{colorTwoThree}{rgb}{0, 1, 1}
\definecolor{colorOneThree}{rgb}{1, 0, 1}
\definecolor{colorOneTwo}{rgb}{1, 1, 0}
\fi

\ifdefined\blackBackground

\else

\fi

\global\long\def\cut#1{}

\global\long\def\reftable#1{table~\ref{#1}}

\global\long\def\detail#1{}

\global\long\def\dataDim{p}

\global\long\def\inputDim{q}

\global\long\def\numData{n}

\global\long\def\numInducing{m}

\global\long\def\dataStd{\sigma}

\global\long\def\dataScalar{y}
\global\long\def\dataVector{\mathbf{\dataScalar}}

\global\long\def\latentScalar{x}
\global\long\def\latentMatrix{\mathbf{\MakeUppercase{\latentScalar}}}
\global\long\def\latentVector{\mathbf{\latentScalar}}

\global\long\def\noiseScalar{\epsilon}

\global\long\def\kernelScalar{k}

\global\long\def\kernelMatrix{\mathbf{\MakeUppercase{\kernelScalar}}}

\global\long\def\numData{n}

\global\long\def\dataDim{p}

\global\long\def\kernel{\kernelScalar}

\global\long\def\meanScalar{\mu}

\global\long\def\mappingFunction{f}
\global\long\def\mappingFunctionVector{\mathbf{\mappingFunction}}

\global\long\def\inducingScalar{u}
\global\long\def\inducingVector{\mathbf{\inducingScalar}}

\global\long\def\inducingInputScalar{z}
\global\long\def\inducingInputVector{\mathbf{\inducingInputScalar}}
\global\long\def\inducingInputMatrix{\mathbf{\MakeUppercase{\inducingInputScalar}}}

\global\long\def\gaussianSamp#1#2{\mathcal{N}\left(#1,#2\right)}
\global\long\def\gaussianDist#1#2#3{\mathcal{N}\left(#1|#2,#3\right)}

\global\long\def\KL#1#2{\text{KL}\left( #1\,\|\,#2 \right)}

\global\long\def\Kff{\kernelMatrix_{\mappingFunctionVector \mappingFunctionVector}}
\global\long\def\Kuu{\kernelMatrix_{\inducingVector \inducingVector}}
\global\long\def\Kuui{\Kuu^{-1}}

\global\long\def\Kfu{\kernelMatrix_{\mappingFunctionVector \inducingVector}}
\global\long\def\Kuf{\kernelMatrix_{\inducingVector \mappingFunctionVector}}
\global\long\def\Qff{{\bf Q}_{\mappingFunctionVector \mappingFunctionVector}}

\newcommand{\latentFuncOne}{f}
\newcommand{\latentFuncTwo}{g}
\newcommand{\fV}{{\bf \latentFuncOne}}
\newcommand{\gV}{{\bf \latentFuncTwo}}
\newcommand{\y}{\dataScalar}
\newcommand{\yV}{\dataVector}
\newcommand{\xV}{\latentVector}
\newcommand{\xM}{\latentMatrix}
\newcommand{\zV}{\inducingInputVector}
\newcommand{\zM}{\inducingInputMatrix}
\newcommand{\muVf}{{\bm \mu}_{\latentFuncOne}}
\newcommand{\muVg}{{\bm \mu}_{\latentFuncTwo}}
\newcommand{\Sf}{{\bm S}_{\latentFuncOne}}
\newcommand{\Sg}{{\bm S}_{\latentFuncTwo}}
\newcommand{\mVf}{{{\bm m}_{\latentFuncOne}}}
\newcommand{\mVg}{{{\bm m}_{\latentFuncTwo}}}
\newcommand{\vVf}{{{\bm v}_{\latentFuncOne}}}
\newcommand{\vVg}{{{\bm v}_{\latentFuncTwo}}}

\newcommand{\fFunc}{f}
\newcommand{\gFunc}{g}
\newcommand{\fX}{\fFunc(\xV)}
\newcommand{\gX}{\gFunc(\xV)}
\newcommand{\uS}{u}
\newcommand{\uV}{{\bf \uS}}
\newcommand{\uVf}{\uV_{\latentFuncOne}}
\newcommand{\uVg}{\uV_{\latentFuncTwo}}
\newcommand{\E}[2]{{\mathbb E}_{#1}\Big[{#2}\Big]}

\newcommand{\Kg}{\kernelScalar_{\latentFuncTwo}}
\newcommand{\Kgg}{\kernelMatrix_{\gV \gV}}
\newcommand{\Kgu}{\kernelMatrix_{\gV \uVg}}
\newcommand{\Kug}{\kernelMatrix_{\uVg \gV}}

\let\oldKff\Kff
\let\oldKfu\Kfu
\let\oldKuf\Kuf
\newcommand{\Kf}{\kernelScalar_{\latentFuncOne}}

\renewcommand*{\Kff}{\kernelMatrix_{\fV \fV}}
\renewcommand*{\Kfu}{\kernelMatrix_{\fV \uVf}}
\renewcommand*{\Kuf}{\kernelMatrix_{\uVf \fV}}

\newcommand{\Kuug}{\kernelMatrix_{\uVg \uVg}} %Fix this notation
\newcommand{\Kuuf}{\kernelMatrix_{\uVf \uVf}} %Fix this notation
\newcommand{\Kuuig}{\kernelMatrix_{\uVg \uVg}^{-1}} %Fix this notation
\newcommand{\Kuuif}{\kernelMatrix_{\uVf \uVf}^{-1}}

\newcommand{\Qgg}{\mathbf{Q}_{\gV \gV}}
\newcommand{\hatQff}{\mathbf{\hat{Q}}_{\fV \fV}}
\newcommand{\hatQgg}{\mathbf{\hat{Q}}_{\gV \gV}}

\newcommand{\numLatentFuncs}{b}
\newcommand{\tTime}{t}
\newcommand{\tTimeV}{\mathbf{\tTime}}
\newcommand{\rTime}{T}
\newcommand{\rTimeV}{\mathbf{\rTime}}
\newcommand{\alphaV}{\mathbf{\alpha}}
\newcommand{\betaV}{\mathbf{\beta}}
\newcommand{\censor}{\delta}
\newcommand{\censorV}{\mathbf{\censor}}
\def\app#1#2{%
  \mathrel{%
    \setbox0=\hbox{$#1\sim$}%
    \setbox2=\hbox{%
      \rlap{\hbox{$#1\propto$}}%
      \lower1.1\ht0\box0%
    }%
    \raise0.25\ht2\box2%
  }%
}

\firstpageno{1}

\makeatletter
\def\@starteditor{\noindent \small {}}
\makeatother

\begin{document}

\title{Chained Gaussian Processes}
\author{\name Alan D. Saul \email alan.saul@sheffield.ac.uk\\
        \addr Department of Computer Science\\
          University of Sheffield
        \AND
        \name James Hensman \email james.hensman@lancaster.ac.uk\\
        \addr CHICAS, Faculty of Health and Medicine\\
              Lancaster University
        \AND
        \name Aki Vehtari \email aki.vehtari@aalto.fi\\
        \addr Helsinki Institute for Information Technology HIIT\\
              Department of Computer Science\\
              Aalto University
        \AND 
        \name Neil D. Lawrence \email n.lawrence@sheffield.ac.uk\\
        \addr Department of Computer Science\\
              University of Sheffield}
\editor{}

\maketitle
\thispagestyle{plain}

\begin{abstract}
Gaussian process models are flexible, Bayesian non-parametric
approaches to regression. Properties of multivariate Gaussians mean that they
can be combined linearly in the manner of additive models and via a link
function (like in generalized linear models) to handle non-Gaussian data.
However, the link function formalism is restrictive, link functions are always
invertible and must convert a parameter of interest to a linear combination of
the underlying processes. There are many likelihoods and models where a
non-linear combination is more appropriate. We term these more general models
\emph{Chained Gaussian Processes}: the transformation of the GPs to the
likelihood parameters will not generally be invertible, and that implies that
linearisation would only be possible with multiple (localized) links, i.e. a
chain. We develop an approximate inference procedure for Chained GPs that is
scalable and applicable to any factorized likelihood. We demonstrate the
approximation on a range of likelihood functions.
\end{abstract}

\section{Introduction}

Gaussian process models are flexible distributions that can provide
priors over non linear functions. They rely on properties of the
multivariate Gaussian for their tractability and their non-parametric
nature. In particular, the sum of two functions, drawn from a Gaussian
process is also given by a Gaussian process. Mathematically, if
$f\sim\gaussianSamp{\meanScalar_f}{\kernelScalar_f}$ and
$g\sim\gaussianSamp{\meanScalar_g}{\kernelScalar_f}$ and we define $y
= g + f$ then properties of multivariate Gaussian give us that $y \sim
\gaussianSamp{\meanScalar_f + \meanScalar_g}{\kernelScalar_f +
  \kernelScalar_g}$ where $\meanScalar_f$ and $\meanScalar_g$ are
deterministic functions of a single input, $\kernelScalar_f$ and
$\kernelScalar_g$ are deterministic, positive semi definite functions
of two inputs and $y$, $g$ and $f$ are stochastic processes.

This elementary property of the Gaussian process is the foundation of much of its power. It makes additive models trivial, and means we can easily combine any process with Gaussian noise. Naturally, it can be applied recursively, and covariance functions can be designed to reflect the underlying structure of the problem at hand (e.g. \citet{Hensman:hierarchical13}~uses additive structure to account for variation in replicate behavior in gene expression).

In practice observations are often \emph{non Gaussian}. In response, statistics has developed the field of \emph{generalized linear models} \citep{Nelder:glm72}. In a generalized linear model a link function is used to connect the mean function of the Gaussian process with the mean function of another distribution of interest. For example, the log link can be used to relate the rate in a Poisson distribution with our GP, $\log \lambda = f + g$. Or for classification the logistic distribution can be used to represent the mean probability of positive outcome, $\log \frac{p}{1-p} = f + g$. 

Writing models in terms of the link function captures the linear nature of the underlying model, but it is somewhat against the probabilistic approach to modeling where we consider the \emph{generative model} of our data. While there's nothing wrong with this relationship mathematically, when we consider the generative model we never apply the link function directly, we consider the inverse link or \emph{transformation function}. For the log link this turns into the exponential,  $\lambda = \exp(f + g)$. Writing the model in this form emphasizes the importance that the transformation function has on the generative model (see e.g.\ work on warped GPs \citep{Snelson:warped04}). The log link implies a multiplicative combination of $f$ and $g$, $\lambda = \exp(f+g) = \exp(f)\exp(g)$, but in some cases we might wish to consider an additive model, $\lambda = \exp(f) + \exp(g)$. Such a model no longer falls within the class of generalized linear models as there is no link function that renders the two underlying processes additive Gaussian. In this paper we address this issue and use variational approximations to develop a framework for \emph{non-linear combination} of the latent processes. Because these models cannot be written in the form a single link function we call this approach  ``Chained Gaussian Processes''.

In this paper we are interested in performing variational inference when we have input-dependent likelihood parameters. We will focus on the case when where the likelihood contains two latent parameters, though the model is general enough to handle more. Parameters of interest could be a latent mean which we wish to infer, a shape parameter for determining the shape of the tails, amongst other things. We will focus on the cases where we have two such latent parameters but propose methods for extending this further.

We will focus on likelihoods $p(\yV|\fV, \gV)$ that depend on two latent functions, $\fV \sim \gaussianDist{\fV}{\muVf}{\fX}$, $\gV \sim \gaussianDist{\gV}{\muVg}{\gX}$. If this noise distribution is a Gaussian then we have two special cases, with $\gX = \dataStd^{2}$ we have a Gaussian process with the conjugate homogeneous Gaussian likelihood. With $\gX = e^{\latentFuncTwo(\xV)}$ we obtain a model for a heteroscedastic Gaussian process~\citep{LazaroGredilla:hetero11}.

A range of other noise models require multiple parameters to be learnt. Traditionally within the Gaussian process literature MAP solutions are used, or alternatively these parameters are integrated out approximately \citep{Rue:CCD09}. In this work we accept that these parameters may change as a function of the input space, and look at inferring posterior Gaussian process functions for these parameters. We do so in a scalable way with sparse variational methods, with the capability of using stochastic gradients during inference. We believe scalability is essential as parameters may only become well determined as the number of observations grow large. We show results with a number of different noise models

We will first introduce notation, and review previous work on heteroscedastic Gaussian processes. Then, we show how to elegantly extend this idea into a more scalable and general framework, allowing a huge number of likelihoods to utilise multiple input dependent processes.

\section{Background}

Assume we have access to a training dataset of $\numData$ input-output observations $\{(\xV_{i}, \y_{i})\}^{\numData}_{i=1}$, $\y_{i}$ is assumed to be a noisy realisation of an underlying latent function $\fV = \fFunc(\xV)$, i.e.\ $\y_{i} = \fFunc(\xV_{i}) + \noiseScalar_{i}$. For a Gaussian likelihood  $\noiseScalar_{i} \sim \gaussianSamp{\mu}{\dataStd^{2}}$, $\xV_{i} \in \mathbb{R}^{\inputDim}$ and $\y_{i} \in \mathbb{R}$. Normally the mean of the likelihood is assumed to be input dependent and given a GP prior $\mu = \fV_{i} = \fFunc(\xV_{i})$ where $\fFunc(\xV) \sim \mathcal{GP}(\muVf, \Kg(\xV, \xV^{\prime}))$, and $\dataStd$ is fixed at an optimal point. In this case the integrals required to infer a posterior, $p(\fV|\yV)$, are tractable.

One extension of this model is the heteroscedastic GP regression model \citep{Bishop:gpsnips97,LazaroGredilla:hetero11}, where the noise variance $\dataStd$ is dependent on the input. The noise variance can be assigned a log GP prior, $\y_{i} \sim \gaussianSamp{\fFunc(\xV_{i})}{e^{\gFunc(\xV_{i})}}$, where $\gFunc(\xV) = \mathcal{GP}(\muVg, \Kg(\xV, \xV^{\prime}))$, i.e.\ a log link function is used. Unfortunately this generalization of the original Gaussian process model is not analytically tractable and requires an approximation to be made. Suggested approximations include MCMC \citep{Bishop:gpsnips97}, variational inference~\citep{LazaroGredilla:hetero11}, Laplace approximation~\citep{Vanhatalo:gpstuff15} and expectation propagation~\citep{Lobato:hetero14} (EP).

Another generalization of the standard GP is to vary the scale of the process as a function of the inputs. \citet{Adams:hetero08} suggest a log GP prior for the scale of the process giving rise to non-parametric non-stationarity in the model. \citet{Turner:pad11} took a related approach to develop probabilistic amplitude demodulation, here the amplitude (or scale) of the process was given by a Gaussian process with a link function given by $\sigma = \log(\exp(f)-1)$. Finally~\citet{Tolvanen:hetero14} assign both the noise variance and the scale a log GP prior.

Both these two variations on Gaussian process regression combine processes in a non-linear way within a Gaussian likelihood, but the idea may be further generalized to systems that deal with non-Gaussian observation noise.

In this paper we describe a general approach to combining processes in a non-linear way. We assume that the likelihood factorizes across the data, but is a general non-linear function of $\numLatentFuncs$ input dependent latent functions. Our main focus will be examples of likelihoods with $\numLatentFuncs = 2$, $\fFunc(\xV_{i})$ and $\gFunc(\xV_{i})$, such that, $p(\yV|\fFunc(\xV_{i}), \gFunc(\xV_{i}))$, though the ideas can all be generalized to $\numLatentFuncs > 2$.
Previous work in this domain include the use of the Laplace approximation~\citep{Vanhatalo:gpstuff15}, however this method scales poorly, $\mathcal{O}(\numLatentFuncs^3\numData^{3})$ and so isn't applicable to datasets of a moderate size. %% It would be possible to extend this model with the inducing point framework, however training would still be impractical for large datasets, without the ability to perform stochastic inference. Additionally~\citet{Lobato:hetero14} perform inference with EP but focus on a heteroscedastic classification model. Our ideas apply to a more general class of likelihoods and also improve on the scalability of this method.
%To our knowledge the only previous work in this domain has focussed on heteroscedastic classification models with inference being performed with EP~\cite{Lobato:hetero14}. Our ideas apply to a more general class of likelihoods and also improve the scalability. The laplace approximation has been used for approximating the posterior 
%Multiple latent functions in with non-Gaussian likelihoods have previously been addressed through the use of a Laplace approximation~\cite{Vanhatalo:gpstuff15}, however the approximation does not scale well, $\mathcal{O}(\numLatentFuncs\numData^{3})$, with the number of data or latent functions.

To render the model tractable we extend recent advances in large scale variational inference approaches to GPs~\citep{Hensman:bigdata13}.  With non-Gaussian likelihoods restrictions on the latent function values may differ, and a non-linear transformation of the latent function, $\gV \in \mathbb{R}^{\inputDim}$ may be required. The inference approach builds on work by~\citet{Hensman:class15}, that in turn builds on the variational inference method proposed by~\citet{Opper:variational09}.

In other work \citep{Nguyen:blackbox14} mixtures of Gaussian latent functions have also been applied for non-Gaussian likelihoods, we expect such mixture distributions would also be applicable to our case. More recently this approach~\citep{Dezfouli:blackbox15} was extended to provide scalability utilising sparse methods similar to this work.

\section{Chained Gaussian Processes}

Our approach to approximate inference in chained GPs builds on previous work in inducing point methods for sparse approximations of GPs~\citep{Snelson:pseudo05,Titsias:variational09,Hensman:class15,Hensman:bigdata13}. Inducing point methods introduce $\numInducing$ `pseudo inputs', known as inducing inputs, at locations $\zM = \{\zV_{i}\}^{\numInducing}_{i=1}$. The corresponding function values are given by $\uV_{i} = \latentFuncOne(\zV_{i})$. These inducing inputs points do not effect the marginal of $\fV$ because
\begin{equation*}
    p(\fV|\xM,\zM) = \int p(\fV|\uV,\xM)p(\uV|\zM)d\uV,
\end{equation*}
where $p(\uV|\zM) = \gaussianDist{\uV}{0}{\Kuu}$ and $p(\fV|\uV,\xM) = \gaussianDist{\fV}{\oldKfu\Kuui\uV}{\oldKff - \oldKfu\Kuui\oldKuf}$. The part-covariances given by $\oldKfu = \kernel_{\latentFuncOne}(\xM,\zM)$ where $\xM$ is the locations of $\fV$, define the relationship between inducing variables and the latent function of interest, $\latentFuncOne$. The marginal likelihood is $p(\yV) = \int p(\yV|\fV)p(\fV|\uV)p(\uV)d\fV\,d\uV$. To avoid $\mathcal{O}(\numData^{3})$ computation complexity~\citet{Titsias:variational09} invokes Jensen's inequality to obtain a lower bound on the marginal likelihood $\log p(\yV)$, an approach known as \emph{variational compression}. This approximation also forms the basis of our approach for non-Gaussian models.

\subsection{Variational Bound}

For non-Gaussian likelihoods, even with a single latent process the marginal likelihood, $p(\yV)$, is not tractable, but it can be lower bounded variationally. We assume that the latent functions, $\fV = \latentFuncOne(\xV)$ and $\gV = \latentFuncTwo(\xV)$ are \emph{a priori} independent
\begin{equation}
    p(\fV,\gV|\uVf,\uVg) = p(\fV|\uVf)p(\gV|\uVg). \label{eq:factFuncs}
\end{equation}
The derivation of the variational lower bound then follows a similar form as~\citep{Hensman:class15} with the extension to multiple latent functions. We begin by writing down our log marginal likelihood,
\begin{align*}
    \log p(\yV) &= \log \int p(\yV|\fV,\gV)p(\fV,\gV|\uVf,\uVg)p(\uVf)p(\uVg) d\fV\,d\gV\,d\uVf\,d\uVg
\end{align*}
then introduce a variational approximation to the posterior,
\begin{equation}
    p(\fV,\gV,\uVg,\uVf|\yV) \approx p(\fV|\uVf)p(\gV|\uVg)q(\uVf)q(\uVg), \label{eq:approxPosterior}
\end{equation}
where we have made the additional assumption that the latent functions factorize in the variational posterior.

Using Jensen's inequality and the factorization of the latent functions~\eqref{eq:factFuncs}, a variational lower bound can then be obtained for the log marginal likelihood,
\begin{align}
    %\log p(\yV) &= \log \int p(\yV|\fV,\gV)p(\fV|\uVf)p(\gV|\uVg)p(\uVf)p(\uVg)\frac{q(\uVf)q(\uVg)}{q(\uVf)q(\uVg)}d\fV\,d\gV\,d\uVf\,d\uVg \nonumber\\
    %&\geq \int q(\uVf)q(\uVg)p(\fV|\uVf)p(\gV|\uVg)\log p(\yV|\fV,\gV) d\fV\,d\gV\,d\uVf\,d\uVg \nonumber
    %\\&\quad - \KL{q(\uVf)}{p(\uVf)} - \KL{q(\uVg)}{p(\uVg)} \nonumber\\
    %&= \int q(\fV)q(\gV)\log p(\yV|\fV,\gV) d\fV\,d\gV - \KL{q(\uVf)}{p(\uVf)} - \KL{q(\uVg)}{p(\uVg)}, \label{eq:bound}
    \log p(\yV) &= \log \int p(\yV|\fV,\gV)p(\fV|\uVf)p(\gV|\uVg)p(\uVf)p(\uVg)d\fV\,d\gV\,d\uVf\,d\uVg \nonumber\\
                &\geq \int q(\fV)q(\gV)\log p(\yV|\fV,\gV) d\fV\,d\gV - \KL{q(\uVf)}{p(\uVf)} - \KL{q(\uVg)}{p(\uVg)}, \label{eq:bound}
\end{align}
where $q(\fV) = \int p(\fV|\uVf)q(\uVf)d\uVf$ and $q(\gV) = \int p(\gV|\uVg)q(\uVg)d\uVg$, and $\KL{p(a)}{p(b)}$ denotes the KL divergence between the two distributions. 
For Gaussian process priors on the latent functions we recover 
\begin{align*}
    p(\fV|\uVf) &= \gaussianDist{\fV}{\Kfu\Kuuif\uVf}{\Kff - \Qff}\\
    p(\gV|\uVg) &= \gaussianDist{\gV}{\Kgu\Kuuig\uVg}{\Kgg - \Qgg},
\end{align*}
where \begin{gather*}\Qff = \Kfu\Kuuif\Kuf \\ \Qgg = \Kgu\Kuuig\Kug. \end{gather*}
Note that covariances for $\fV$ and $\gV$, can differ though their inducing input locations, $\zM$, are shared.

We take $q(\uVf)$ and $q(\uVg)$ to be Gaussian distributions with variational parameters, ${q(\uVf) = \gaussianDist{\uVf}{\muVf}{\Sf}}$ and ${q(\uVg) = \gaussianDist{\uVg}{\muVg}{\Sg}}$. Using the properties of multivariate Gaussians this results in tractable integrals for $q(\fV)$ and $q(\gV)$,
\begin{align}
    q(\fV) &= \gaussianDist{\fV}{\Kfu\Kuuif\muVf}{\Kff + \hatQff}\label{eq:qf}\\
    q(\gV) &= \gaussianDist{\gV}{\Kgu\Kuuig\muVg}{\Kgg + \hatQgg}\label{eq:qg},
\end{align}
where \begin{gather*} \hatQff = \Kfu\Kuuif(\Sf - \Kuuf)\Kuuif\Kuf \\ \hatQgg = \Kgu\Kuuig(\Sg - \Kuug)\Kuuig\Kug. \end{gather*}
%Need to work out what these are when the mean is non zero

The KL terms in~\eqref{eq:bound} and their derivative can be computed in closed form and are inexpensive as they are divergence between Gaussians. However, an intractable integral, ${\int q(\fV)q(\gV)\log p(\yV|\fV,\gV) d\fV\,d\gV}$, still remains.
Since the likelihood factorizes,
\begin{equation*}
    p(\yV|\fV,\gV) = \prod\nolimits^{\numData}_{i=1} p(\yV_{i}|\fV_{i},\gV_{i}),
\end{equation*}
the problematic integral in~\eqref{eq:bound} also factorizes across data points, allowing us to use stochastic variational inference~\citep{Hensman:bigdata13,Hoffman:stochastic13},
\begin{align}
    \int q(\fV)q(\gV)\log p(\yV|\fV,\gV) d\fV\,d\gV &= \int q(\fV)q(\gV)\log \prod^{\numData}_{i=1}p(\yV_{i}|\fV_{i},\gV_{i}) d\fV\,d\gV \nonumber\\
    &= \sum\nolimits^{\numData}_{i=1} \int q(\fV_{i})q(\gV_{i})\log p(\yV_{i}|\fV_{i},\gV_{i}) d\fV_{i}\,d\gV_{i}.\label{eq:keyint}
\end{align}
We are then left with $\numData$, $\numLatentFuncs$ dimensional Gaussian integrals over the log-likelihood,
\begin{align}
    \log p(\yV) &\geq \sum\nolimits^{\numData}_{i=1} \int q(\fV_{i})q(\gV_{i}) \log p(\yV_{i}|\fV_{i},\gV_{i})d\fV_{i}\,d\gV_{i}\nonumber\\
                &\quad- \KL{q(\uVf)}{p(\uVf)} - \KL{q(\uVg)}{p(\uVg)}.\label{eq:finalbound}
\end{align}
The bound will also hold for any additional number of latent functions
by assuming they all factorize in the variational
posterior. 
%Figure~\ref{fig:graphicalmodel} shows a graphical model of
%the model prior to integration of $\fV$ and $\gV$.  

%\begin{figure}
  %\centering
  %\resizebox{0.3\textwidth}{!}{%
  %\tikz{%
    %\node[obs] (x) {$x$} ; %
    %\node[latent, above of=x,yshift=25] (uf) {$\uVf$} ; %
    %\node[latent, below of=x,yshift=-25] (ug) {$\uVg$} ; %
    %\node[latent, right=1 of uf] (f) {$\fV_{i}$} ; %
    %\node[latent, right=1 of ug] (g) {$\gV_{i}$} ; %
    %\edge[->] {x} {f} ; 
    %\edge[->] {x} {g} ;
    %\edge[->] {uf} {f} ; 
    %\edge[->] {ug} {g} ;
    %\node[obs, right=1 of f, yshift=-25] (y) {$\yV_{i}$} ; %
    %\edge[->] {f} {y} ;
    %\edge[->] {g} {y} ;
    %\edge[-] {f} {f} ; %circular
    %\edge[-] {g} {g} ; %circular
    %\plate {fgy} {(f)(y)(g)} {$\forall i \in \numData$}
  %}
  %}
  %\caption{Graphical model describing the chained GP model. In the posterior note $\fV$ and $\gV$ are integrated out analytically.}
  %\label{fig:graphicalmodel}
%\end{figure}

The bound decomposes into a sum over data, as such the $\numData$ input points
can be visited in mini-batches, and the gradients and log-likelihood
of each mini-batch can be subsequently summed, this operation can be
also be parallelized~\citep{Gal:distributed14}. A single mini-batch can
instead be visited obtaining a stochastic gradient for use in a
stochastic
optimization~\citep{Hensman:bigdata13,Hoffman:stochastic13}. This
provides the ability to scale to huge datasets.

If the likelihood is Gaussian these integrals are analytic \citep{LazaroGredilla:hetero11}, though the noise variance must be constrained positive via a transformation of the latent function, e.g\ an exponent. In this case,
\begin{align*}
    &\int q(\fV_{i})q(\gV_{i}) \log p(\yV_{i}|\fV_{i},\gV_{i})d\fV_{i}\,d\gV_{i}\\
    &\quad= \int \gaussianDist{\fV_{i}}{\mVf_i}{\vVf_i}\gaussianDist{\gV_{i}}{\mVg_i}{\vVg_i} \log \gaussianDist{\yV_{i}}{\fV_{i}}{e^{\gV_i}}\\
    &\quad= \log \gaussianDist{\yV_{i}}{\mVf_i}{e^{\mVg_i - \frac{\vVg_i}{2}}} - \frac{\vVg_i}{4} - \frac{\vVf_ie^{-\mVg_i + \frac{\vVg_i}{2}}}{2}
\end{align*}
where we define \begin{align*}\hfill & \mVf = \Kfu\Kuuif\muVf & \hfill & \vVf = \Kff + \hatQff \\ \hfill & \mVg = \Kgu\Kuuig\muVg & \hfill & \vVg = \Kgg + \hatQgg. \end{align*} $\vVf_i$ denotes the $i$th diagonal element of the matrix with $\vVf$ along its diagonal.
It may be possible in this Gaussian case to find the optimal $q(\fV)$ such that the bound collapses to that of~\citet{LazaroGredilla:hetero11}, however this would not allow for stochastic optimization. Here we arrive at a sparse extension, where a Gaussian distribution is assumed for the posterior over of $\fV$, where as previously $q(\fV)$ has been collapsed out and could take any form. This sparse extension provides the ability to scale to much larger datasets whilst maintaining a similar variational lower bound.

%%{{\fixmem{
%Possibly mention that you can do it similarly to Aki with the signal and noise variance = 3 latent functions?
%Maybe do the student-T version instead?
%Explain how the non-zero mean function matters
%%}}

The model is however not restricted to heteroscedastic Gaussian likelihoods. If the integral~\eqref{eq:keyint} and its gradients can be computed in an unbiased way, any factorizing likelihood can be used.
This can be seen as a chained Gaussian process. There is no single link function that allows the specification of this model under the modelling assumptions of a generalised linear model.
An example that will be revisited in the experiments is the beta distribution, $\y_{i} \sim B(\alpha, \beta)$ where $\alpha, \beta \in \mathbb{R}^{+}$ and observations $\y_{i} \in (0, 1)$, $\xV_{i} \in \mathbb{R}^{\inputDim}$. Since $\alpha, \beta$ must maintain positiveness, then can be assigned log GP priors, 
\begin{equation}
    \y_{i} \sim B(\alpha = e^{\fFunc(\xV_{i})}, \beta = e^{\gFunc(\xV_{i})}),\label{eq:beta}
\end{equation}
where $\fFunc(\xV) = \mathcal{GP}(\muVf, \Kf(\xV, \xV^{\prime}))$ and $\gFunc(\xV) = \mathcal{GP}(\muVg, \Kg(\xV, \xV^{\prime}))$. This allows the shape of the beta distribution to change over time, see Supplementary Material and section~\ref{sec:twitter} for an example plots.

%% To our knowledge such a model hasn't previously been proposed, however 
Using the variational bound above, all that is required is to perform a series of $\numData$ two dimensional quadratures, for both the log-likelihood and its gradients, a relatively simple task and computationally feasible when looking at modest batch sizes. From this example the power and adaptability of the method should be apparent. 

A major strength of this method is that performing this integral is the only requirement to implement a new noise model, similarly to~\citep{Nguyen:blackbox14,Hensman:class15}. Further, since a stochastic optimizer is used the gradients do not need to be exact. Our implementations can use off the shelf stochastic optimizer, such as Adagrad~\citep{Duchi:adagrad11} or RMSProp~\citep{Tieleman:RMSProp12}. Further, for many likelihoods some portion of these integrals is analytically tractable, reducing the variance introduced by numerical integration. See supplementary material for an investigation. %For higher number of latent functions higher dimensional quadrature would be required. %In these cases it may be more efficient to make use of low variance Monte Carlo estimates for the integrals.% In the next section we compare Monte Carlo and quadrature for the two dimensional case.

\subsection{Posterior and Predictive Distributions}

Following from~\eqref{eq:approxPosterior} it is clear that when the variational lower bound bound has been maximised with respect to the variational parameters, $p(\uVf|\yV) \approx q(\uVf)$ and $p(\uVg|\yV) \approx q(\uVg)$. The posterior for $p(\fV^{*}|\yV^{*})$ under this approximation is
\begin{align*}
    p(\fV^{*}|\yV^{*}) &= \int p(\fV^{*}|\xV, \fV)p(\fV|\uVf)p(\uVf|\yV)d\fV\,d\uVf\\
    &= \int p(\fV^{*}|\uVf)p(\uVf|\yV)d\uVf\\
    &\approx \int p(\fV^{*}|\uVf)q(\uVf)d\uVf = q(\fV^{*}),
\end{align*}
where $q(\fV^{*})$ and $q(\gV^{*})$ become similar to~\eqref{eq:qf}.
%\begin{align*}
    %q(\fV^{*}) &= \gaussianDist{\fV^{*}}{\Kfupred\Kuuif\muVf}{\Kffpred + \Kfupred\Kuuif(\Sf - \Kff)\Kuuif\Kufpred}\\
    %q(\gV^{*}) &= \gaussianDist{\gV^{*}}{\Kgupred\Kuuif\muVf}{\Kggpred + \Kgupred\Kuuif(\Sf - \Kff)\Kuuif\Kugpred}
%\end{align*}

Finally, treating each prediction point independently, the predictive distribution for each data pair $\{(\xV_{i}^{*}, \yV^{*}_{i})\}^{\numData^{*}}_{i=1}$ follows as
\begin{align*}
    p(\yV_{i}^{*}|\yV_{i}, \xV_{i}) &= \int p(\yV_{i}^{*}|\fV_{i}^{*},\gV_{i}^{*})q(\fV_{i}^{*})q(\gV_{i}^{*}) d\fV_{i}^{*}\,d\gV_{i}^{*}.
\end{align*}
This integral is analytically intractable in the general case, but again can be computed using a series of two dimensional quadrature or simple Monte Carlo sampling.

\section{Experiments}

To evaluate the effectiveness of our chained GP approximations we consider a range of real and synthetic datasets. The performance measure used throughout is the negative log predictive density (NLPD) on held out data, \reftable{tab:results}.\footnote{Data used in the experiments can be downloaded via the pods package: \url{https://github.com/sods/ods}} The results for mean absolute error (MAE) (Supplementary Material) show comparable results between methods. 5-fold cross-validation is used throughout. The non-linear optimization of (hyper-) parameters is subject to local minima, as such multiple runs were performed on each fold with a range of parameter initialisations. The solution obtaining the highest log-likelihood on the training data of each fold was retained. Automatic relevance determination exponentiated quadratic kernels are used throughout allowing one lengthscale per input dimension, in addition to a bias kernel.\footnote{Code is publically available at: \url{https://github.com/SheffieldML/ChainedGP}}.. In all experiments 100 inducing points were used and their locations were optimized with respect to the lower bound of the log marginal likelihood following~\citep{Titsias:variational09}.

\begin{table*}%[t!]
    \centering
    {\small
    \begin{tabular}{c c c c c c}
        \hline
        \multicolumn{1}{c}{\textbf{Data}} & \multicolumn{4}{c}{\textbf{NLPD}}\\
        \multicolumn{1}{c}{} & \multicolumn{1}{c}{\textbf{G}} & \multicolumn{1}{c}{\textbf{CHG}} &\multicolumn{1}{c}{\textbf{Lt}} &\multicolumn{1}{c}{\textbf{Vt}} & \multicolumn{1}{c}{\textbf{CHt}}\\
        \hline
elevators1000  &  $0.39 \pm 0.13$ &   $0.1 \pm 0.01$ &               NA &               NA &               NA \\
elevators10000 &  $0.07 \pm 0.01$ &  $0.03 \pm 0.02$ &               NA &               NA &               NA \\
motorCorrupt   &  $2.04 \pm 0.06$ &  $1.79 \pm 0.05$ &  $1.73 \pm 0.05$ &  $2.52 \pm 0.09$ &   $1.7 \pm 0.05$ \\
boston         &  $0.27 \pm 0.02$ &  $0.09 \pm 0.01$ &  $0.23 \pm 0.02$ &  $0.19 \pm 0.02$ &  $0.09 \pm 0.02$ \\
        %old
        %Elevators1000 & 0.295 & 0.230 & - & - \\ %This is based on the three folds that worked for both models (initialization issues) at the minute, will change when experiments are rerun
        %Elevators10000 & 0.197 & 0.111 & - & - \\ 
        %CorruptMotor & 1.832 & 1.637 & 1.593 & 1.563 \\ %Corrupted silverman motorcycle dataset, with extra noisy elements added
        %Boston & 0.321 & 0.115 & 0.214 & 0.103 \\
        \hline
    \end{tabular}
    \caption{Results NLPD over 5 cross-validation folds with 10 replicates each. Models shown in comparison are sparse Gaussian (G), chained heteroscedastic Gaussian (CHG), Student-$t$ Laplace approximation (Lt), Student-$t$ VB approximation (Vt), and chained heteroscedastic Student-$t$ (CHt).}
    \label{tab:results}
    %\vspace{-1em}
}
\end{table*}
\subsection{Heteroscedastic Gaussian}

In our introduction we related our ideas to heteroscedastic GPs. We will first use our approximations to show how the addition of input dependent noise to a Gaussian process regression model effects performance, compared with a sparse Gaussian process model~\citep{Titsias:variational09}. Performance is shown to improve as more data is provided as would be expected, making it clear that both models can scale with data, though the new model is more flexible when handling the distributions tails. A sparse Gaussian process with Gaussian likelihood is chosen in these experiments as a baseline, as a non-sparse Gaussian process cannot scale to the size of all the experiments. 

The Elevator1000 uses a subset of $1,000$ of the Elevator dataset
%\footnote{Dataset available at \url{http://www.liaad.up.pt/~ltorgo/Regression/DataSets.html}}
. In this data the heteroscedastic model (Chained GP) offers considerable improvement in terms of negative log predictive density (NLPD) over the sparse GP (Table~\ref{tab:results}.
%, MAE error results show that the models are comparable. This is in correspondence with many of our experiments that show most of the improvement from the introduction of heterogeneity come being able to explain the data in the tails more accurately. 
Our second experiment with the Gaussian likelihood, Elevator10000, examines scaling of the model. Here a subset of $10,000$ data points of the Elevator dataset are used, and performance is improved as expected. Previous models for heteroscedastic Gaussian process models cannot scale, the chained GP can implement the heteroscedastic setting and scale. %Experiments for the following model were excluded as the dataset is not subject to outliers\fixmem{?}.
%model standard variational sparse Gaussian process when both are assuming a Gaussian likelihood, and then how performance can be improved by the capability of utilising more data.
%Following existing work in heteroschedastic GPs~\cite{LazaroGredilla:hetero11,Kersting:hetero07} first show that our bound is tight and predictive performance comparable on small datasets.

%{\fixmem{
    %\begin{enumerate}
        %\item Goldberg Synthetic, 100, 1000, then 100000 for HSVGP
        %\item Yuan and Wahba dataset has non-linear function for g(x)
        %%\item Motorcycle dataset - maybe check out student-t performance aswell? Whats the point? It just shows for 100 data that we can do heteroschedastic
        %\item Elevators, 1000 training, 3000 testing, then 10,000 training, 3000 testing
    %\end{enumerate}
%}
%}
\begin{figure}
    \centering
    \begin{minipage}{.33\textwidth}
        \resizebox{\textwidth}{!}{%
            % This file was created by matplotlib v0.1.0.
% Copyright (c) 2010--2014, Nico Schlömer <nico.schloemer@gmail.com>
% All rights reserved.
% 
% The lastest updates can be retrieved from
% 
% https://github.com/nschloe/matplotlib2tikz
% 
% where you can also submit bug reports and leavecomments.
% 
\begin{tikzpicture}

\definecolor{color1}{rgb}{0.892228375410333,0.594078451097012,0.476491355597973}
\definecolor{color0}{rgb}{0.454117654263973,0.706666679680347,0.62705884128809}
\definecolor{color3}{rgb}{0.848925037839833,0.59604231550413,0.75221147190122}
\definecolor{color2}{rgb}{0.590569805257461,0.641520974846447,0.759710894752951}
\definecolor{color4}{rgb}{0.633906976194943,0.769267214186051,0.406096126846239}

\begin{axis}[
xmin=-0.5, xmax=4.5,
ymin=0.5, ymax=4.5,
xtick={0,1,2,3,4},
xticklabels={G,Lt,CHG,CHt,Vt},
ymajorgrids
]
\addplot [white!34.82353001832962!black]
coordinates {
(2.22044604925031e-16,1.80882934016417)
(2.22044604925031e-16,1.62650101035581)

};
\addplot [white!34.82353001832962!black]
coordinates {
(2.22044604925031e-16,2.13531302121023)
(2.22044604925031e-16,2.61871941435328)

};
\addplot [white!34.82353001832962!black]
coordinates {
(-0.2,1.62650101035581)
(0.2,1.62650101035581)

};
\addplot [white!34.82353001832962!black]
coordinates {
(-0.2,2.61871941435328)
(0.2,2.61871941435328)

};
\addplot [white!34.82353001832962!black]
coordinates {
(-0.4,1.87995259959198)
(0.4,1.87995259959198)

};
%\addplot [line width=0.7000000000000001pt, white!34.82353001832962!black, mark=diamond, mark size=1]
%coordinates {
%(2.22044604925031e-16,3.04032939263803)
%(2.22044604925031e-16,3.51529459350683)
%(2.22044604925031e-16,3.14362236093065)
%(2.22044604925031e-16,2.76662592034076)

%};
\addplot [line width=0.7000000000000001pt, white!34.82353001832962!black, mark=diamond, mark size=1]
coordinates {
(2.22044604925031e-16,3.04032939263803)
};
\addplot [line width=0.7000000000000001pt, white!34.82353001832962!black, mark=diamond, mark size=1]
coordinates {
(2.22044604925031e-16,3.51529459350683)
};
\addplot [line width=0.7000000000000001pt, white!34.82353001832962!black, mark=diamond, mark size=1]
coordinates {
(2.22044604925031e-16,3.14362236093065)
};
\addplot [line width=0.7000000000000001pt, white!34.82353001832962!black, mark=diamond, mark size=1]
coordinates {
(2.22044604925031e-16,2.76662592034076)
};
\addplot [white!34.82353001832962!black]
coordinates {
(1,1.55943053868191)
(1,1.22143100320048)

};
\addplot [white!34.82353001832962!black]
coordinates {
(1,1.82267149023705)
(1,2.20135153320343)

};
\addplot [white!34.82353001832962!black]
coordinates {
(0.8,1.22143100320048)
(1.2,1.22143100320048)

};
\addplot [white!34.82353001832962!black]
coordinates {
(0.8,2.20135153320343)
(1.2,2.20135153320343)

};
\addplot [white!34.82353001832962!black]
coordinates {
(0.6,1.70317697132861)
(1.4,1.70317697132861)

};
%\addplot [line width=0.7000000000000001pt, white!34.82353001832962!black, mark=diamond, mark size=1]
%coordinates {
%(1,1.15828108358106)
%(1,0.878835244030063)
%(1,2.2680336573943)
%(1,2.24780817752013)
%(1,2.3160146814781)
%(1,2.35710262018191)
%(1,2.79159936727861)
%};
\addplot [line width=0.7000000000000001pt, white!34.82353001832962!black, mark=diamond, mark size=1]
coordinates {
(1,1.15828108358106)
};
\addplot [line width=0.7000000000000001pt, white!34.82353001832962!black, mark=diamond, mark size=1]
coordinates {
(1,0.878835244030063)
};
\addplot [line width=0.7000000000000001pt, white!34.82353001832962!black, mark=diamond, mark size=1]
coordinates {
(1,2.2680336573943)
};
\addplot [line width=0.7000000000000001pt, white!34.82353001832962!black, mark=diamond, mark size=1]
coordinates {
(1,2.24780817752013)
};
\addplot [line width=0.7000000000000001pt, white!34.82353001832962!black, mark=diamond, mark size=1]
coordinates {
(1,2.3160146814781)
};
\addplot [line width=0.7000000000000001pt, white!34.82353001832962!black, mark=diamond, mark size=1]
coordinates {
(1,2.35710262018191)
};
\addplot [line width=0.7000000000000001pt, white!34.82353001832962!black, mark=diamond, mark size=1]
coordinates {
(1,2.79159936727861)

};
\addplot [white!34.82353001832962!black]
coordinates {
(2,1.59529076672889)
(2,1.09831097516445)

};
\addplot [white!34.82353001832962!black]
coordinates {
(2,1.98728960623159)
(2,2.53348455126858)

};
\addplot [white!34.82353001832962!black]
coordinates {
(1.8,1.09831097516445)
(2.2,1.09831097516445)

};
\addplot [white!34.82353001832962!black]
coordinates {
(1.8,2.53348455126858)
(2.2,2.53348455126858)

};
\addplot [white!34.82353001832962!black]
coordinates {
(1.6,1.70162715107047)
(2.4,1.70162715107047)

};
%\addplot [line width=0.7000000000000001pt, white!34.82353001832962!black, mark=diamond, mark size=1]
%coordinates {
%(2,0.88612166771596)
%(2,2.62756801329519)
%(2,2.79883859537933)
%
%};
\addplot [line width=0.7000000000000001pt, white!34.82353001832962!black, mark=diamond, mark size=1]
coordinates {
(2,0.88612166771596)

};
\addplot [line width=0.7000000000000001pt, white!34.82353001832962!black, mark=diamond, mark size=1]
coordinates {
(2,2.62756801329519)

};
\addplot [line width=0.7000000000000001pt, white!34.82353001832962!black, mark=diamond, mark size=1]
coordinates {
(2,2.79883859537933)

};
\addplot [white!34.82353001832962!black]
coordinates {
(3,1.5725473577058)
(3,1.20747654865959)

};
\addplot [white!34.82353001832962!black]
coordinates {
(3,1.82393618792976)
(3,2.19276735601684)

};
\addplot [white!34.82353001832962!black]
coordinates {
(2.8,1.20747654865959)
(3.2,1.20747654865959)

};
\addplot [white!34.82353001832962!black]
coordinates {
(2.8,2.19276735601684)
(3.2,2.19276735601684)

};
\addplot [white!34.82353001832962!black]
coordinates {
(2.6,1.68063221363743)
(3.4,1.68063221363743)

};
%\addplot [line width=0.7000000000000001pt, white!34.82353001832962!black, mark=diamond, mark size=1]
%coordinates {
%(3,1.04693047167048)
%(3,0.789486201272578)
%(3,2.24210221266637)
%(3,2.31643368295767)
%(3,2.2527027257266)
%(3,2.73821086772194)
%
%};
\addplot [line width=0.7000000000000001pt, white!34.82353001832962!black, mark=diamond, mark size=1]
coordinates {
(3,1.04693047167048)
};
\addplot [line width=0.7000000000000001pt, white!34.82353001832962!black, mark=diamond, mark size=1]
coordinates {
(3,0.789486201272578)
};
\addplot [line width=0.7000000000000001pt, white!34.82353001832962!black, mark=diamond, mark size=1]
coordinates {
(3,2.24210221266637)
};
\addplot [line width=0.7000000000000001pt, white!34.82353001832962!black, mark=diamond, mark size=1]
coordinates {
(3,2.31643368295767)
};
\addplot [line width=0.7000000000000001pt, white!34.82353001832962!black, mark=diamond, mark size=1]
coordinates {
(3,2.2527027257266)
};
\addplot [line width=0.7000000000000001pt, white!34.82353001832962!black, mark=diamond, mark size=1]
coordinates {
(3,2.73821086772194)

};
\addplot [white!34.82353001832962!black]
coordinates {
(4,2.17772068941547)
(4,1.41143821839779)

};
\addplot [white!34.82353001832962!black]
coordinates {
(4,2.79958694348952)
(4,3.6856595301203)

};
\addplot [white!34.82353001832962!black]
coordinates {
(3.8,1.41143821839779)
(4.2,1.41143821839779)

};
\addplot [white!34.82353001832962!black]
coordinates {
(3.8,3.6856595301203)
(4.2,3.6856595301203)

};
\addplot [white!34.82353001832962!black]
coordinates {
(3.6,2.55359355275534)
(4.4,2.55359355275534)

};
%\addplot [line width=0.7000000000000001pt, white!34.82353001832962!black, mark=diamond, mark size=1]
%coordinates {
%(4,1.13843126636047)
%(4,3.83595009019722)
%(4,3.814044834044)
%(4,4.34233049091983)
%
%};
\addplot [line width=0.7000000000000001pt, white!34.82353001832962!black, mark=diamond, mark size=1]
coordinates {
(4,1.13843126636047)
};
\addplot [line width=0.7000000000000001pt, white!34.82353001832962!black, mark=diamond, mark size=1]
coordinates {
(4,3.83595009019722)
};
\addplot [line width=0.7000000000000001pt, white!34.82353001832962!black, mark=diamond, mark size=1]
coordinates {
(4,3.814044834044)
};
\addplot [line width=0.7000000000000001pt, white!34.82353001832962!black, mark=diamond, mark size=1]
coordinates {
(4,4.34233049091983)
};
\path [draw=white!34.82353001832962!black, fill=color0] (axis cs:-0.4,1.80882934016417)--(axis cs:0.4,1.80882934016417)--(axis cs:0.4,2.13531302121023)--(axis cs:-0.4,2.13531302121023)--(axis cs:-0.4,1.80882934016417)--cycle;

\path [draw=white!34.82353001832962!black, fill=color1] (axis cs:0.6,1.55943053868191)--(axis cs:1.4,1.55943053868191)--(axis cs:1.4,1.82267149023705)--(axis cs:0.6,1.82267149023705)--(axis cs:0.6,1.55943053868191)--cycle;

\path [draw=white!34.82353001832962!black, fill=color2] (axis cs:1.6,1.59529076672889)--(axis cs:2.4,1.59529076672889)--(axis cs:2.4,1.98728960623159)--(axis cs:1.6,1.98728960623159)--(axis cs:1.6,1.59529076672889)--cycle;

\path [draw=white!34.82353001832962!black, fill=color3] (axis cs:2.6,1.5725473577058)--(axis cs:3.4,1.5725473577058)--(axis cs:3.4,1.82393618792976)--(axis cs:2.6,1.82393618792976)--(axis cs:2.6,1.5725473577058)--cycle;

\path [draw=white!34.82353001832962!black, fill=color4] (axis cs:3.6,2.17772068941547)--(axis cs:4.4,2.17772068941547)--(axis cs:4.4,2.79958694348952)--(axis cs:3.6,2.79958694348952)--(axis cs:3.6,2.17772068941547)--cycle;

\path [draw=white, fill opacity=0] (axis cs:13,4.5)--(axis cs:13,4.5);

\path [draw=white, fill opacity=0] (axis cs:4.5,13)--(axis cs:4.5,13);

\path [draw=white, fill opacity=0] (axis cs:13,0.5)--(axis cs:13,0.5);

\path [draw=white, fill opacity=0] (axis cs:-0.5,13)--(axis cs:-0.5,13);

\end{axis}

\end{tikzpicture}
        }
        \subcaption{\small{NLPD motor}}
        \label{fig:NLPDMotor}
    \end{minipage}
    %\begin{subfigure}[t]{.23\textwidth}
        %\resizebox{\textwidth}{!}{%
            %%\input{corrupt_motorcycle_rel_log_pred.tikz}
            %\input{motorCorrupt_rel_NLPD.tikz}
        %}
        %\caption{\small{Relative NLPD motor}}
        %\label{fig:RNLPDMotor}
    %\end{subfigure}
    \begin{minipage}{.33\textwidth}
        \resizebox{\textwidth}{!}{%
            % This file was created by matplotlib v0.1.0.
% Copyright (c) 2010--2014, Nico Schlömer <nico.schloemer@gmail.com>
% All rights reserved.
% 
% The lastest updates can be retrieved from
% 
% https://github.com/nschloe/matplotlib2tikz
% 
% where you can also submit bug reports and leavecomments.
% 
\begin{tikzpicture}

\definecolor{color1}{rgb}{0.892228375410333,0.594078451097012,0.476491355597973}
\definecolor{color0}{rgb}{0.454117654263973,0.706666679680347,0.62705884128809}
\definecolor{color3}{rgb}{0.848925037839833,0.59604231550413,0.75221147190122}
\definecolor{color2}{rgb}{0.590569805257461,0.641520974846447,0.759710894752951}
\definecolor{color4}{rgb}{0.633906976194943,0.769267214186051,0.406096126846239}

\begin{axis}[
xmin=-0.5, xmax=4.5,
ymin=-0.2, ymax=0.7,
xtick={0,1,2,3,4},
xticklabels={G,Lt,CHG,CHt,Vt},
ymajorgrids
]
\addplot [white!34.82353001832962!black]
coordinates {
(2.22044604925031e-16,0.178294861109533)
(2.22044604925031e-16,0.011396699952241)

};
\addplot [white!34.82353001832962!black]
coordinates {
(2.22044604925031e-16,0.356005954898338)
(2.22044604925031e-16,0.615628627654936)

};
\addplot [white!34.82353001832962!black]
coordinates {
(-0.2,0.011396699952241)
(0.2,0.011396699952241)

};
\addplot [white!34.82353001832962!black]
coordinates {
(-0.2,0.615628627654936)
(0.2,0.615628627654936)

};
\addplot [white!34.82353001832962!black]
coordinates {
(-0.4,0.24498220411849)
(0.4,0.24498220411849)

};
\addplot [line width=0.7000000000000001pt, white!34.82353001832962!black, mark=diamond, mark size=1]
coordinates {

};
\addplot [white!34.82353001832962!black]
coordinates {
(1,0.13123396457477)
(1,0.0042984969227103)

};
\addplot [white!34.82353001832962!black]
coordinates {
(1,0.306362897305006)
(1,0.547039174962776)

};
\addplot [white!34.82353001832962!black]
coordinates {
(0.8,0.0042984969227103)
(1.2,0.0042984969227103)

};
\addplot [white!34.82353001832962!black]
coordinates {
(0.8,0.547039174962776)
(1.2,0.547039174962776)

};
\addplot [white!34.82353001832962!black]
coordinates {
(0.6,0.209387356841222)
(1.4,0.209387356841222)

};
\addplot [line width=0.7000000000000001pt, white!34.82353001832962!black, mark=diamond, mark size=1]
coordinates {
(1,0.56972577284605)

};
\addplot [white!34.82353001832962!black]
coordinates {
(2,0.0151582878963988)
(2,-0.117110158174859)

};
\addplot [white!34.82353001832962!black]
coordinates {
(2,0.139305247649752)
(2,0.26976231629554)

};
\addplot [white!34.82353001832962!black]
coordinates {
(1.8,-0.117110158174859)
(2.2,-0.117110158174859)

};
\addplot [white!34.82353001832962!black]
coordinates {
(1.8,0.26976231629554)
(2.2,0.26976231629554)

};
\addplot [white!34.82353001832962!black]
coordinates {
(1.6,0.0911878003134029)
(2.4,0.0911878003134029)

};
\addplot [line width=0.7000000000000001pt, white!34.82353001832962!black, mark=diamond, mark size=1]
coordinates {
(2,0.32732824902131)
(2,0.330682310602644)

};
\addplot [white!34.82353001832962!black]
coordinates {
(3,0.0115472964121733)
(3,-0.110476724906809)

};
\addplot [white!34.82353001832962!black]
coordinates {
(3,0.163551308653174)
(3,0.307729107376433)

};
\addplot [white!34.82353001832962!black]
coordinates {
(2.8,-0.110476724906809)
(3.2,-0.110476724906809)

};
\addplot [white!34.82353001832962!black]
coordinates {
(2.8,0.307729107376433)
(3.2,0.307729107376433)

};
\addplot [white!34.82353001832962!black]
coordinates {
(2.6,0.0790168410465746)
(3.4,0.0790168410465746)

};
\addplot [line width=0.7000000000000001pt, white!34.82353001832962!black, mark=diamond, mark size=1]
coordinates {
(3,0.399691846183182)

};
\addplot [white!34.82353001832962!black]
coordinates {
(4,0.10097956435031)
(4,-0.023627969603755)

};
\addplot [white!34.82353001832962!black]
coordinates {
(4,0.274604680096325)
(4,0.414245637066389)

};
\addplot [white!34.82353001832962!black]
coordinates {
(3.8,-0.023627969603755)
(4.2,-0.023627969603755)

};
\addplot [white!34.82353001832962!black]
coordinates {
(3.8,0.414245637066389)
(4.2,0.414245637066389)

};
\addplot [white!34.82353001832962!black]
coordinates {
(3.6,0.190415731775828)
(4.4,0.190415731775828)

};
\addplot [line width=0.7000000000000001pt, white!34.82353001832962!black, mark=diamond, mark size=1]
coordinates {

};
\path [draw=white!34.82353001832962!black, fill=color0] (axis cs:-0.4,0.178294861109533)--(axis cs:0.4,0.178294861109533)--(axis cs:0.4,0.356005954898338)--(axis cs:-0.4,0.356005954898338)--(axis cs:-0.4,0.178294861109533)--cycle;

\path [draw=white!34.82353001832962!black, fill=color1] (axis cs:0.6,0.13123396457477)--(axis cs:1.4,0.13123396457477)--(axis cs:1.4,0.306362897305006)--(axis cs:0.6,0.306362897305006)--(axis cs:0.6,0.13123396457477)--cycle;

\path [draw=white!34.82353001832962!black, fill=color2] (axis cs:1.6,0.0151582878963988)--(axis cs:2.4,0.0151582878963988)--(axis cs:2.4,0.139305247649752)--(axis cs:1.6,0.139305247649752)--(axis cs:1.6,0.0151582878963988)--cycle;

\path [draw=white!34.82353001832962!black, fill=color3] (axis cs:2.6,0.0115472964121733)--(axis cs:3.4,0.0115472964121733)--(axis cs:3.4,0.163551308653174)--(axis cs:2.6,0.163551308653174)--(axis cs:2.6,0.0115472964121733)--cycle;

\path [draw=white!34.82353001832962!black, fill=color4] (axis cs:3.6,0.10097956435031)--(axis cs:4.4,0.10097956435031)--(axis cs:4.4,0.274604680096325)--(axis cs:3.6,0.274604680096325)--(axis cs:3.6,0.10097956435031)--cycle;

\path [draw=white, fill opacity=0] (axis cs:13,0.7)--(axis cs:13,0.7);

\path [draw=white, fill opacity=0] (axis cs:4.5,13)--(axis cs:4.5,13);

\path [draw=white, fill opacity=0] (axis cs:13,-0.2)--(axis cs:13,-0.2);

\path [draw=white, fill opacity=0] (axis cs:-0.5,13)--(axis cs:-0.5,13);

\end{axis}

\end{tikzpicture}
        }
        \subcaption{\small{NLPD Boston}}
        \label{fig:NLPDBoston}
    \end{minipage}
    %\begin{subfigure}[t]{.23\textwidth}
        %\resizebox{\textwidth}{!}{%
            %%\input{boston_rel_log_pred.tikz}
            %\input{boston_rel_NLPD.tikz}
        %}
        %\caption{\small{Relative NLPD Boston}}
        %\label{fig:RNLPDBoston}
    %\end{subfigure}
    \caption{\protect\subref{fig:NLPDMotor}) NLPD on corrupt motorcycle dataset. \protect\subref{fig:NLPDBoston}) NLPD of Boston housing dataset. In NLPD lower is better, models shown in comparison are sparse Gaussian (G), Student-$t$ Laplace approximation (Lt), Student-$t$ VB approximation (Vt), chained heteroscedastic Gaussian (CHG), and chained heteroscedastic Student-$t$ (CHt). Boxplots show the variation over 5 folds.}
    \label{fig:results}
\end{figure}
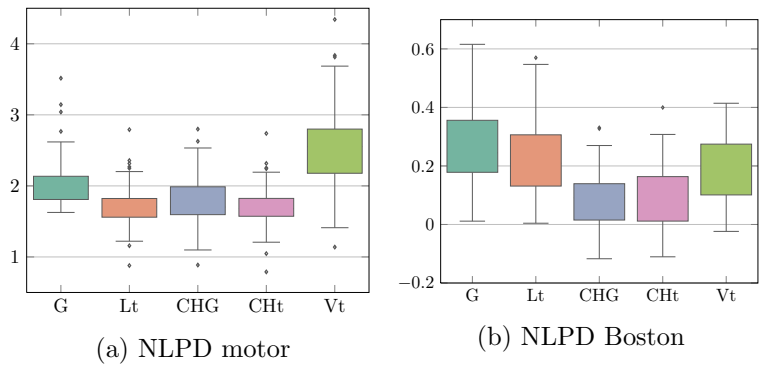

\subsection{Non-Gaussian Heteroscedastic likelihoods}
\label{sec:nonGaussian}

One of the major strengths of the approximation over pure scalability, is the ability to use more general non-Gaussian likelihoods.  In this section we will investigate this flexibility by performing inference with non-standard likelihoods. This allows models to be specified that correspond to the modellers belief about the data in a flexible way.

We first investigate an extension of the Student-$t$ likelihood
that endows it with an input-dependent scale
parameter. This is straightforward in the chained GP
paradigm. 

\begin{figure*}
    \resizebox{1.0\textwidth}{!}{%
        \input{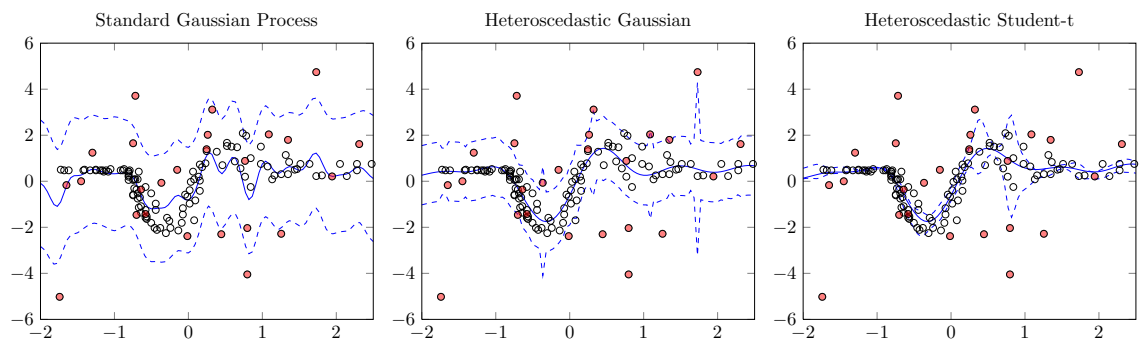}
    }
    \caption{Corrupted motorcycle dataset, fitted with a Gaussian
      process model with a Gaussian likelihood, a Gaussian process
      with input dependent noise (heteroscedastic) with a Gaussian
      likelihood, and a Gaussian process with Student-$t$ likelihood,
      with an input dependent shape parameter. The mean is
      shown in solid and the variance is shown as dotted}
    \label{fig:corruptMotor}
\end{figure*}

The corrupt motorcycle dataset is an artificial modification to the
benchmark motorcycle dataset~\citep{Silverman:splines85} and shows the
models capabilities more clearly. The original motorcycle dataset has
had 25 of its data randomly corrupted with Gaussian noise of
$\gaussianSamp{0}{3}$, simulating spurious accelerometer readings.  
We hope that our method will be robust and ignore such outlying values.  
An input-dependent mean, $\mu$, is set alongside an input dependent scale which
must be positive, $\sigma$. A constant degrees of freedom
parameter $\nu$, is initalized to 4.0 and then is optimized to its MAP solution.
\begin{equation}
    \y_{i} \sim St(\mu = \fFunc(\xV_{i}), \sigma^{2}= e^{\gFunc(\xV_{i})}, \nu) \label{eq:stut}
\end{equation}
where $\fFunc(\xV) = \mathcal{GP}(\muVf, \Kf(\xV, \xV^{\prime}))$ and
$\gFunc(\xV) = \mathcal{GP}(\muVg, \Kg(\xV, \xV^{\prime}))$. This
provides a heteroscedastic extension to the Student-$t$ likelihood. We
compare the model with a Gaussian process with homogeneous Student-$t$
likelihood, approximated variationally~\citep{Hensman:class15} and the Laplace approximation.
Figure~\ref{fig:corruptMotor} shows the improved quality of the error bars with the chained heteroscedastic Student-$t$ model. Learning a model
with heavy tails allows outliers to be ignored, and so its input dependent variance can
be collapsed around just points close to the underlying function,
which in this case is known to be well modelled with a heteroscedastic
Gaussian process~\citep{LazaroGredilla:hetero11,Bishop:gpsnips97}. It
is also interesting to note the heteroscedastic Gaussian's performance,
although not able to completely ignore outliers the model has learnt a
very short lengthscale. This renders the prior over the scale
parameter independent across the data, meaning that the resulting
likelihood is more akin to a scale-mixture of Gaussians (which endows
appropriate robustness characteristics). The main difference is that
the scale-mixture is based on a log-Gaussian prior, as opposed to the
Student-$t$ which is based on an inverse Gamma.

Figure~\ref{fig:results} shows the NLPD on the corrupt motorcycle dataset and Boston
housing dataset. The Boston housing dataset shows the median house
prices throughout the Boston area, quantified by 506 data points, with
13 explanatory input variables~\citep{Kuss:robust06}. We find that the chained heteroscedastic Gaussian process model
already outperforms the Student-$t$ model on this dataset, and the additional ability to use heavier tails in the chained Student-$t$ is not used. This ability to regress back to an already powerful model is a useful property of the chained Student-$t$ model.

\begin{figure*}
    %\begin{subfigure}{0.95\textwidth}
    \includegraphics[width=\textwidth,keepaspectratio]{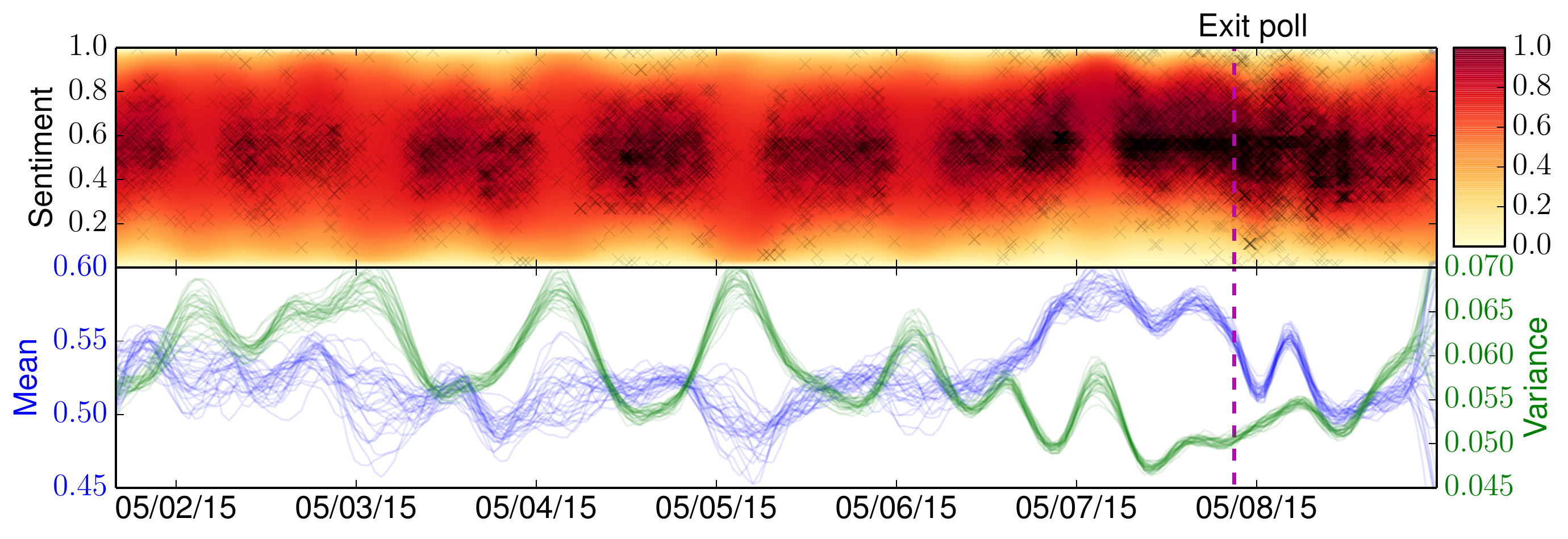}
    \caption{Twitter sentiment from the UK general election modelled using a heteroscedastic beta distribution. The timing of the exit poll is marked and is followed by a night of tweets as election counts come in. Other night time periods have a reduced volume of tweets and a corresponding increase in sentiment variance. Ticks on the x-axis indicate midnight.}
    \label{fig:twitterpdf}
\end{figure*}

%\fixmem{?}What happens when fitting Gaussian data with this model, does it revert back? When fitting Gaussian data with the student-$t$, does df go large?

\subsubsection{Survival Analysis}
Survival analysis focuses on the analysis of time-to-event data. This data arises frequently in clinical trials, though it is also commonly found in failure tests within engineering. In these settings it is common to observe censoring. Censoring occurs when an event is only observed to exist between two times, but no further information is available. For right-censoring, the most common type of censoring, the event time $\rTime \in [t, \infty)$.

A common model to analyse this type of data is an \emph{accelerated failure time} model. This suggests that the distribution of when an random event, $\rTime$, may occur, is multiplicatively effected by some function of the covariates, $f(\xV)$, thus accelerating or retarding time; akin to notion of dog years. In a generalized linear model we may write this as $\log \rTime = \log \rTime_{0} + \log f(\xV)$, where $\rTime$ is the random variable for failure time of the individual with covariates $\xV$, and $\rTime_{0}$ follows a parametric distribution describing a non-accelerated failure time.

\begin{figure}
    \includegraphics[width=\textwidth,keepaspectratio,resolution=300]{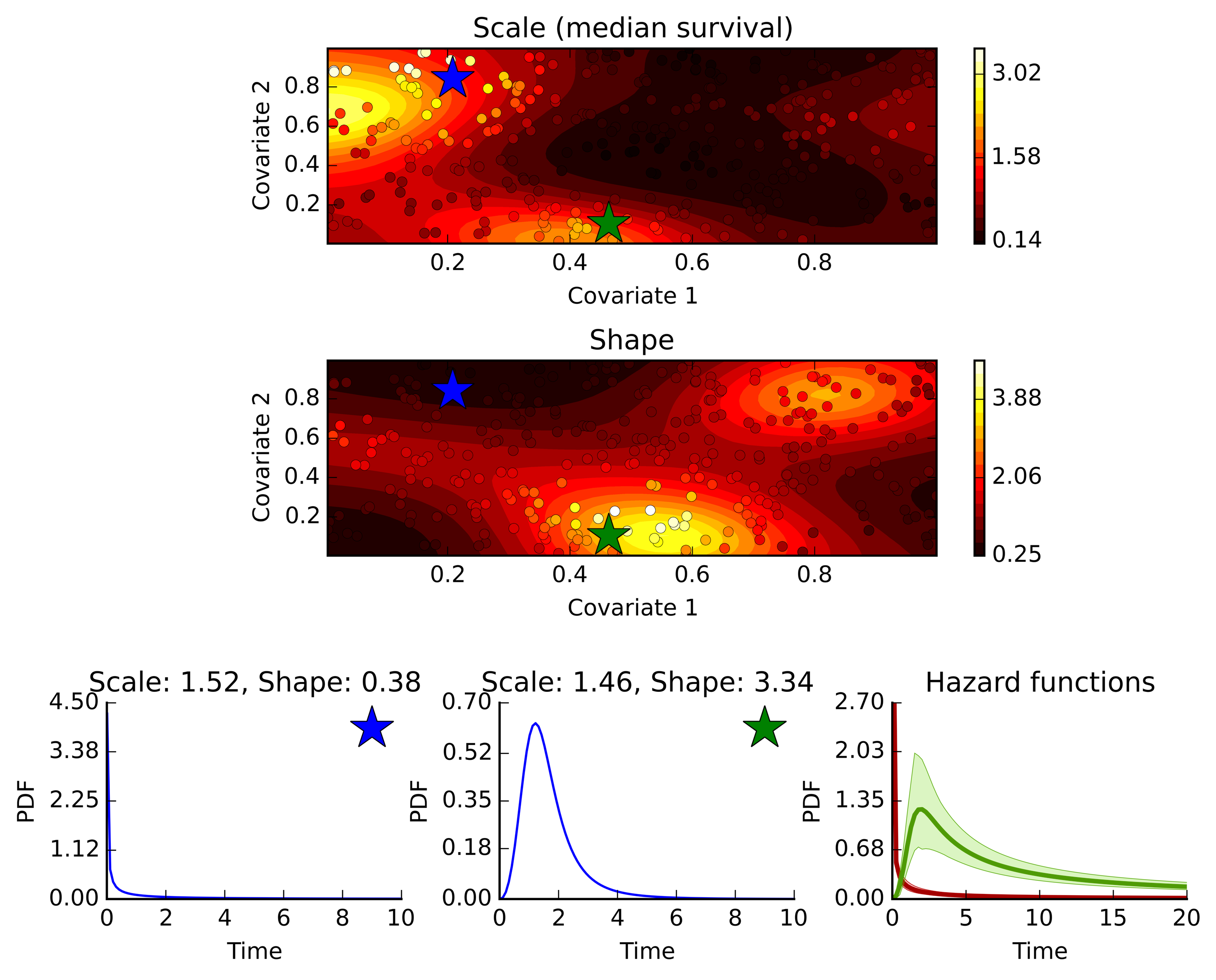}
    \caption{Resulting model on synthetic survival dataset. Shows variation of median survival time and shape of log-logistic distribution, in response to differing covariate information. Background colour shows the chained-survivals predictions, coloured dots show ground truth. Lower figures show associated failure time distributions and hazards for two different synthetic patients. Shapes can be both unimodal or exponential.}
    \label{fig:survivalDist}
\end{figure}

To account for censoring the cumulative distribution needs to be computable and event times are restricted to be positive. A common parametric distribution for $\rTime_{0}$ that fulfills these restrictions is the log-logistic distribution, with the median being some function of the covariates, $f(\xV)$. This however is restrictive as the \emph{shape} of failure time distribution is assumed to be the same for all patients. We relax this assumption by allowing the shape parameter of the log-logistic distribution to vary with response to the input,
\begin{equation*}
    \y_{i} \sim LL(\alpha = e^{\fFunc(\xV_{i})}, \beta = e^{\gFunc(\xV_{i})}),
\end{equation*}
where $\fFunc(\xV) = \mathcal{GP}(\muVf, \Kf(\xV, \xV^{\prime}))$ and $\gFunc(\xV) = \mathcal{GP}(\muVg, \Kg(\xV, \xV^{\prime}))$. This allows both skewed unimodal and exponential shaped distributions for the failure time distribution depending on the individual, as shown in Figure~\ref{fig:survivalDist}. Again there is no associated link-function in this case, and the model can be modelled as a chained-survival model.

\begin{table}
    \centering
    \small{
    \begin{tabular}{c c c c c}
        \hline
        \multicolumn{1}{c}{\textbf{Data}} & \multicolumn{4}{c}{\textbf{NLPD}}\\
        \multicolumn{1}{c}{} & \multicolumn{1}{c}{\textbf{G}} & \multicolumn{1}{c}{\textbf{LSurv}} & \multicolumn{1}{c}{\textbf{VSurv}} & \multicolumn{1}{c}{\textbf{CHSurv}}\\
        \hline
leuk &  $4.03 \pm .08$ &  $1.57 \pm .01$ &  $1.57 \pm .01$ &  $1.56 \pm .01$ \\
Surv &  $5.45 \pm .06$ &  $2.52 \pm .02$ &  $2.52 \pm .02$ &  $2.16 \pm .02$ \\
        \hline
    \end{tabular}
    \caption{Results NLPD over 5 cross-validation folds with 10 replicates each. Models shown in comparison are sparse Gaussian (G), survival Laplace approximation (LSurv), survival VB approximation (VSurv), chained heteroscedastic survival (CHSurv).}
    \label{tab:survival}
    %\vspace{-1em}
}
\end{table}

Table~\ref{tab:survival} shows the models performance a real and synthetic datasets. The leukemia dataset~\citep{henderson:leukemia2002} contains censored event times for 1043 leukemia patients and is known to have non-linear responses certain covariates~\citep{Gelman:BDA32013}. We find little advantage from using the chained-survival model, but as usual the model is robust such that performance isn't degraded in this case. We additionally show the results on a synthetic dataset where the shape parameter is known to vary with response to the input, in this case an increase in performance is seen. See Appendix~\ref{appendix:survival} for more details on the model and synthetic dataset.

%This suggests that the 'hazard' of an event occuring at any time, is a product of a baseline hazard$, \lambda_{0}$, and some function of the covariates, $f(\xV)$, giving $\lambda (\tTimeV|\xV) = f(\xV)\lambda_{0}(f(\xV)\tTimeV)$.

\subsubsection{Twitter Sentiment Analysis in the UK Election}
\label{sec:twitter}

The final experiment shows the adaptability of the model even further, on a novel dataset and with a novel heteroscedastic model. We consider sentiment in the UK general election, focussing on tweets tagged as supporters of the Labour party. We used a sentiment analysis tagging system\footnote{Available from \url{https://www.twinword.com/}} to evaluate the positiveness of 105,396 tweets containing hashtags relating to recent the major political parties, over the run in to the UK 2015 general election. 

We are interested in modeling the distribution of positive sentiment as a function of time. The sentiment value is constrained to be to be between zero and one, and we do not believe the distribution of tweets through time to be necessarily unimodal. A natural likelihood to use in this case is the beta likelihood. This allows us to accommodate bathtub shaped distributions, indicating tweets are either extremely positive or extremely negative. We then allow the distribution over tweets to be heterogenous throughout time by using Gaussian process models for each parameter of the beta distribution, 
\begin{equation*}
    \y_{i} \sim B(\alpha = e^{\fFunc(\xV_{i})}, \beta = e^{\gFunc(\xV_{i})}),
\end{equation*} 
where $\fFunc(\xV) = \mathcal{GP}(\muVf, \Kf(\xV, \xV^{\prime}))$ and $\gFunc(\xV) = \mathcal{GP}(\muVg, \Kg(\xV, \xV^{\prime}))$. 

The upper section of Figure~\ref{fig:twitterpdf} shows the data and the probability of each sentiment value throughout time. The lower part shows the corresponding mean and variance functions induced by the above parameterization. This year's general election was particularly interesting: polls throughout the election showed it to be a close race between the two major parties, Conservative and Labour. But at the end of polling an exit poll was released that predicted an outright win for the Conservatives. This exit poll proved accurate and is associated with a corresponding dip in the sentiment of the tweets. Other interesting aspects of the analysis include the reduction in number of tweets during the night and the corresponding increase in the variance of our estimates. 

\subsubsection{Decomposition of Poisson Processes}

The intensity, $\lambda(x)$, of a Poisson process can be modelled as the product of two positive latent functions, $\exp(f(x))$ and $\exp(g(x))$, as a generalrised linear model,
\begin{gather*}
    \log(\lambda) = f(x) + g(x)\\
    y \sim \textrm{Poisson}(\lambda = \exp(f+g) = \exp(f(x))\exp(g(x))), 
\end{gather*}
using a \emph{log} link function.

Instead imagine we form a new process by combining two different underlying Poisson processes through addition. The superposition property of Poissons means that the resulting process is also Poisson with rates given by the sum of the underlying rates.

To model this via a Gaussian process we have to assume that the intensity of the resulting Poisson, $\lambda(x)$ is a \emph{sum} of two positive functions, which are denoted by $\exp(f(x))$ and $\exp(g(x))$ respectively,
\begin{align}
    y \sim \textrm{Poisson}(\lambda = \exp(f(x)) + \exp(g(x))),
    \label{eq:multipoisson}
\end{align}
there is no link function representation for this model, it takes the form of a chained-GP.

\begin{figure}
    \centering
    \includegraphics[trim={0 0.7cm 0 0},clip,width=0.7\textwidth]{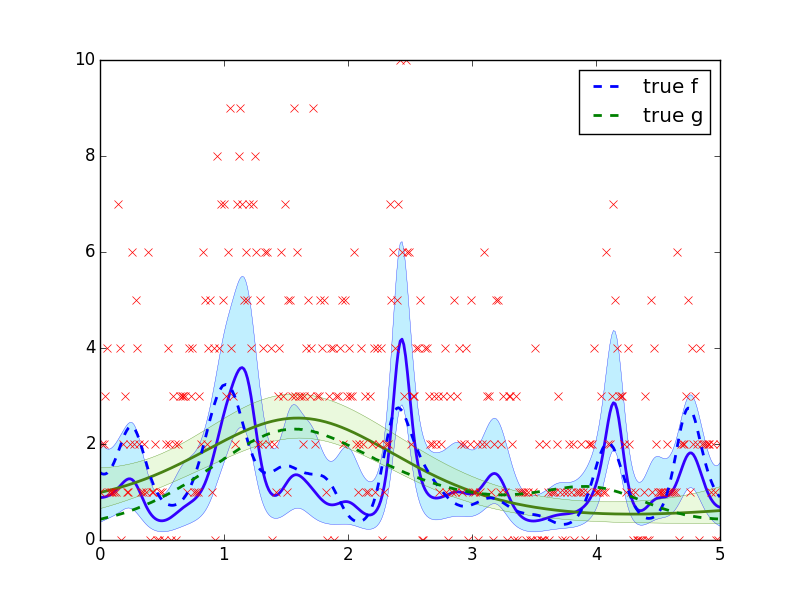}
    \caption{Even with 350 data we can start to see the differentiation of the addition of a long lengthscale positive process and a short lengthscale positive process. Red crosses denote observations, dotted lines are the true latent functions generating the data using Eq~\eqref{eq:multipoisson}, the solid line and associated error bars are the approximate posterior predictions, $q(\fV^{*}), q(\gV^{*})$, of the latent processes.}
    \label{fig:multipoisson}
\end{figure}

Focusing purely on the generative model of the data, the lack of an link function does not present an issue. Figure~\ref{fig:multipoisson} shows a simple demonstration of the idea in a simulated data set. 

Using an additive model for the rate rather than a multiplicative model for counting processes has been discussed previously in the context of linear models for survival analysis, with promising results~\cite{lin:additive1995}.

\begin{figure}
\begin{center}
    \includegraphics[width=0.43\textwidth,keepaspectratio]{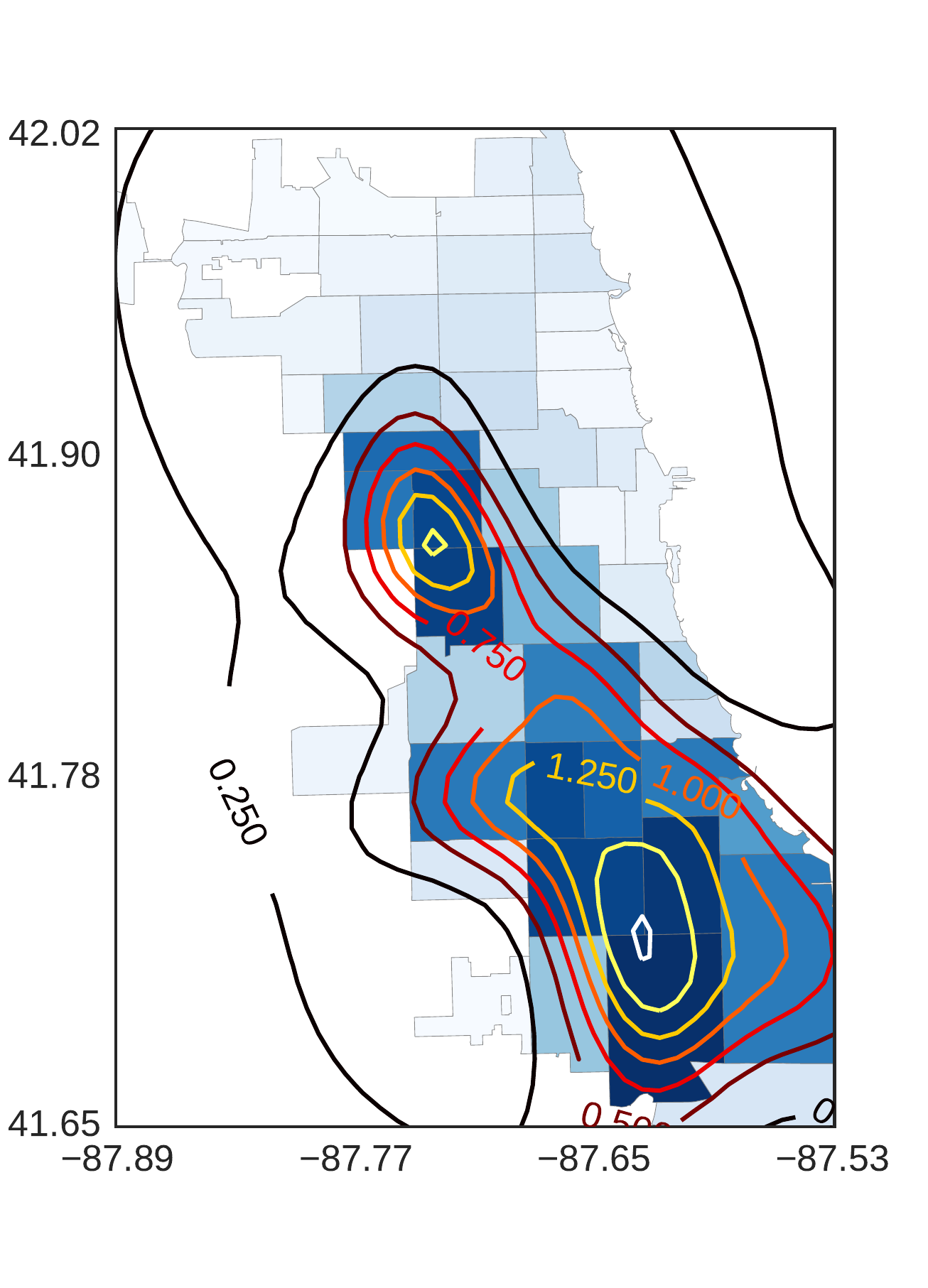}
    \includegraphics[width=0.43\textwidth,keepaspectratio]{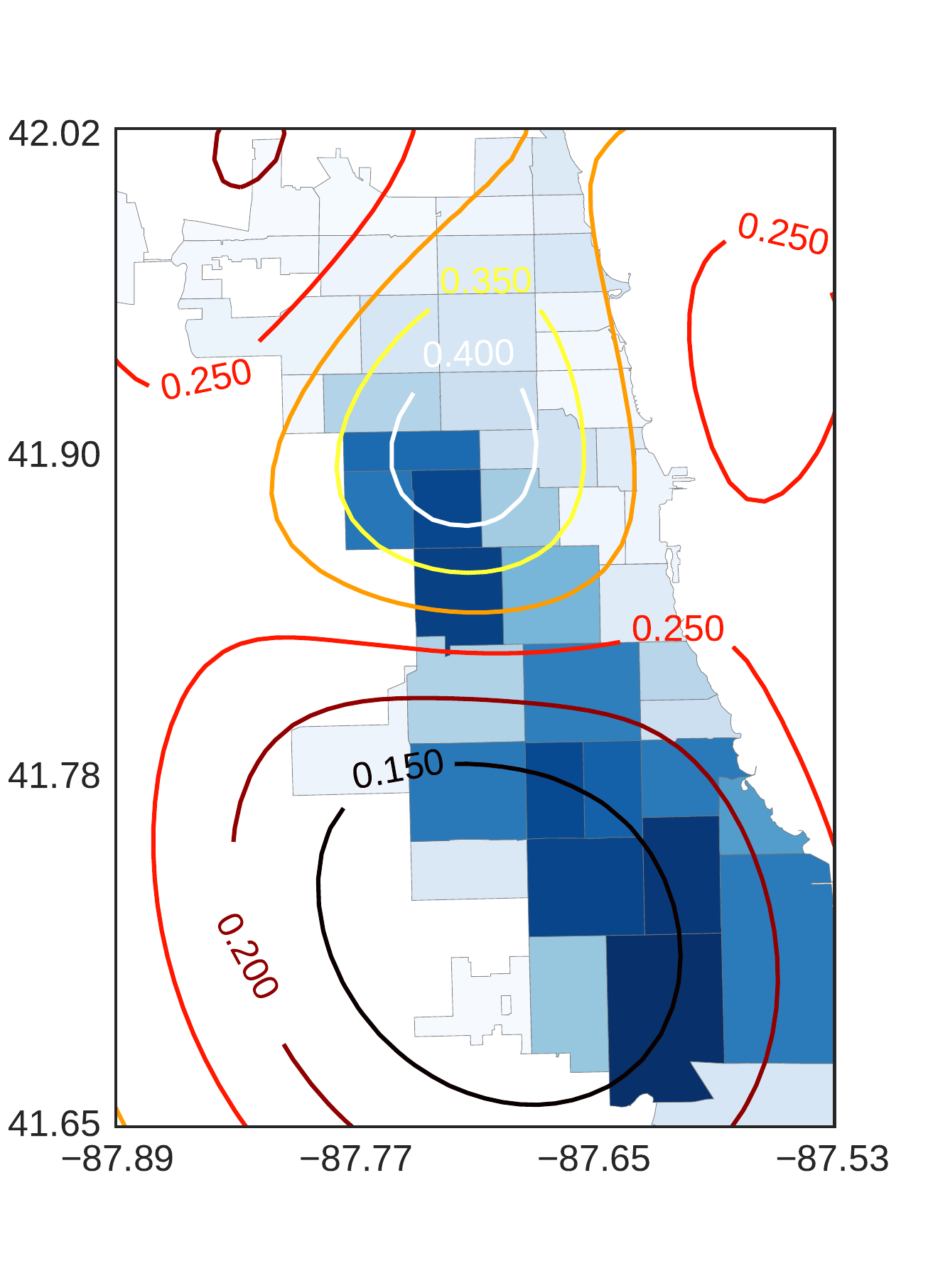}\\
    \includegraphics[width=0.43\textwidth,keepaspectratio]{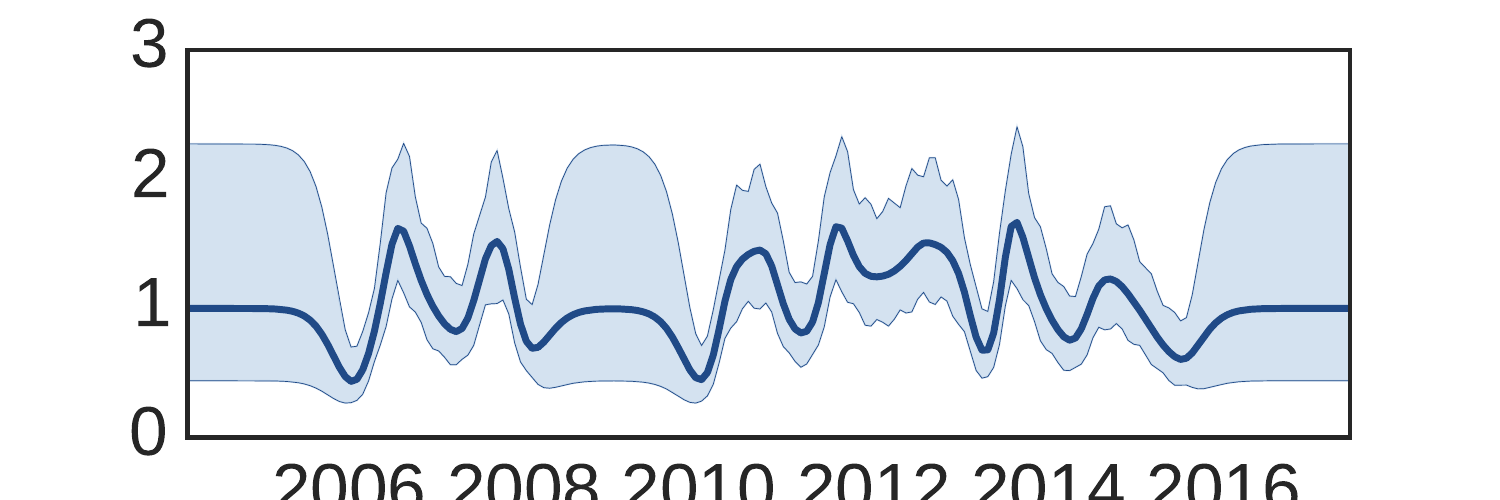}
    \includegraphics[width=0.43\textwidth,keepaspectratio]{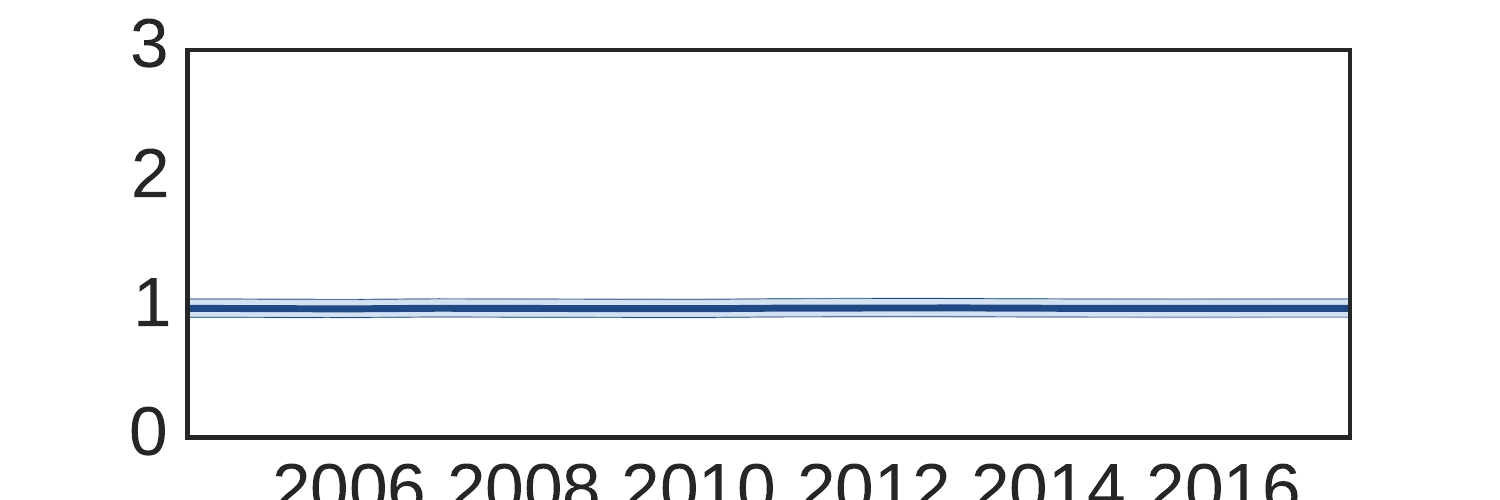}
    \caption{Homicide rate maps for Chicago. The short length scale spatial process, $\lambda_1(x)$ (above-left) is multiplied in the model by a temporal process, $\mu_1(t)$ (below-left) which fluctuates with passing seasons. Contours of spatial process are plotted as deaths per month per zip code area. Error bars on temporal processes are at 5th and 95th percentile. The longer length scale spatial process, $\lambda_2(x)$ (above-right) has been modeled with little to no fluctuation temporally $\mu_2(t)$ (below-right). }\label{fig:short_length_chicago}
\end{center}
\end{figure}

To illustrate the model on real data we considered homicide data in Chicago. Taking data from \url{http://homicides.redeyechicago.com/}~(see also \citet{adams:point2014}) we aggregated data into three months periods by zip code. We considered an additive Poisson process with a particular structure for the covariance functions. We constructed a rate of the form:
\[
\Lambda(x, t) = \lambda_1(x)\mu_1(t) + \lambda_2(x)\mu_2(t)
\]
where $\lambda_1(x)=\exp(f_1(x))$, $\lambda_2(x) = \exp(g_1(x))$, $\mu_1(t)=\exp(f_2(t))$ and $\mu_2(t)=\exp(g_2(t))$ where $f_1(x)$, $g_1(x)$ are spatial GPs and $f_2(t)$ and $g_2(t)$ are temporal GPs. The overall rate decomposes into two separable rate functions, but the overall rate function is not separable. We have a sum of separable \citep{Alvarez:vector12} rate functions. This structure allows us to decompose the homicide map into separate spatial maps that each evolve at different time rates. We selected one spatial map with a length scale of 0.04 and one spatial map with a length scale of 0.09. The time scales and variances of the temporal rate functions were optimized by maximum likelihood. The results are shown in Figure \ref{fig:short_length_chicago}. The long length scale process hardly fluctuates across time, whereas the short lengthscale process, which represents more localized homicide activity, fluctuates across the seasons with scaled increases of around 1.25 deaths per month per zip code. This decomposition is possible and interpretable due to the structured underlying nature of the GPs inside the chained model.

%\section{TODO}
%Gaussian
%Heteroscedastic - M. Lazaro-Gredilla and M.K. Titsias (2011). Variational Heteroscedastic Gaussian Process Regression.
%datasets - http://www.dcc.fc.up.pt/~ltorgo/Regression/DataSets.html
%RMSE and LPD
%\begin{enumerate}
    %\item Elevators, with 3000 to compare, and more than 3000 to show the power
    %\item Abalone, Pole Telecommunications, larger N
%\end{enumerate}
%StudentT
%MAE and LPD, explain why MAE and why LPD is more appropriate
%Compare to Robust Regression with Twinned Gaussian Process, run on Friedman with small number of N, then much larger.
%\begin{enumerate}
    %\item Friedman
    %\item Boston Housing
%\end{enumerate}
%Beta distribution, 
%RMSE and LPD vs SVI-GP (which should do badly)
%Plots
%\begin{enumerate}
    %\item Twitter dataset
%\end{enumerate}
%Survival data
%Laplace with log-logistic fixed $\beta$ parameter
%Compare to?
%\begin{enumerate}
    %\item Leukemia
    %\item Dream?
    %\item Artificial larger N
%\end{enumerate}
%MCMC comparisons?
%Results table

\section{Conclusions}

We have introduced ``Chained Gaussian Process'' models. They allow us to make predictions which are based on a non-linear combination of underlying latent functions. This gives a far more flexible formalism than the generalized linear models that are classically applied in this domain.

Chained Gaussian processes are a general formalism and therefore are intractable in the base case. We derived an approximation framework that is applicable for any factorized likelihood. For the cases we considered, involving two latent functions, the approximation made use of two dimensional Gauss-Hermite quadrature. We speculated that when the idea is extended to higher numbers of latent functions it may be necessary to resort to Monte Carlo sampling.

Our approximation is highly scalable through the use of stochastic variational inference. This enables the full range of standard stochastic optimizers to be applied in the framework.

\subsubsection*{Acknowledgments}

AS was supported by a University of Sheffield, Faculty Scholarship, JH was supported by a MRC fellowship. The authors also thank Amazon for a donation of AWS compute time and the anonymous reviewers of a previous transcript of this work.

{\small
\bibliography{../../../bib/lawrence,../../../bib/other,../../../bib/zbooks,../../bib/library}
}

\appendix
\clearpage

\section{Supplementary Material}
%\end{subfigure}
%\begin{subfigure}{.49\textwidth}
    %\resizebox{\textwidth}{!}{%
        %\input{labour_latent.tikz}
    %}
%\end{subfigure}
%\begin{subfigure}{.49\textwidth}
    %\resizebox{\textwidth}{!}{%
        %\input{labour_mean.tikz}
    %}
%\end{subfigure}
%\begin{subfigure}{.49\textwidth}
    %\resizebox{\textwidth}{!}{%
        %\input{labour_var.tikz}
    %}
%\end{subfigure}
%%\resizebox{.5\textwidth}{!}{%
%%\input{labour_modes.tikz}
%%}

\subsection{Collapsed Heteroscedastic}
\citet{LazaroGredilla:hetero11} form a bound by `collapsing out' the $q(\fV)$ distribution, such that it need not take a Gaussian form.
As a brief review their bound can be derived as follows.
\begin{align*}
    \log p(\yV|\fV) &\geq \E{q(\fV)}{\log p(\yV|\fV, \gV)} + \log \E{q(\gV)}{\frac{p(\gV)}{q(\gV)}}\\
    &= \log \gaussianDist{\yV}{\fV}{e^{\mVf_{i} - \frac{\vVf_{i}}{2}}} - \frac{1}{4}\sum^{n}_{i=1}\vVf_{i} = L^\prime \\
\end{align*}
\begin{align*}
    L &= \int p(\yV|\fV)p(\fV)d\fV\\
      &\geq \int e^{L^{\prime}}p(\fV)d\fV\\
      &= \int \gaussianDist{\yV}{\fV}{e^{\mVf_{i} - \frac{\vVf_{i}}{2}}}\gaussianDist{\fV}{0}{\Kff} - \frac{1}{4}\sum^{n}_{i=1}\vVf_{i} \\&\quad- \KL{q(\gV)}{p(\gV)}\\
      &= \gaussianDist{\yV}{0}{\Kff + e^{\mVf_{i} - \frac{\vVf_{i}}{2}}} - \frac{1}{4}\sum^{n}_{i=1}\vVf_{i} \\&\quad- \KL{q(\gV)}{p(\gV)}\\
\end{align*}
The bound that we assumes a sparse approximation, however it also constrains $q(\fV)$ to be Gaussian. This leads to an additional KL divergence since the optimal is not chosen, and additional penalty term arising from the mismatch of the constrained form of $q(\fV)$.

\subsection{Quadrature and Monte Carlo}

\begin{figure}
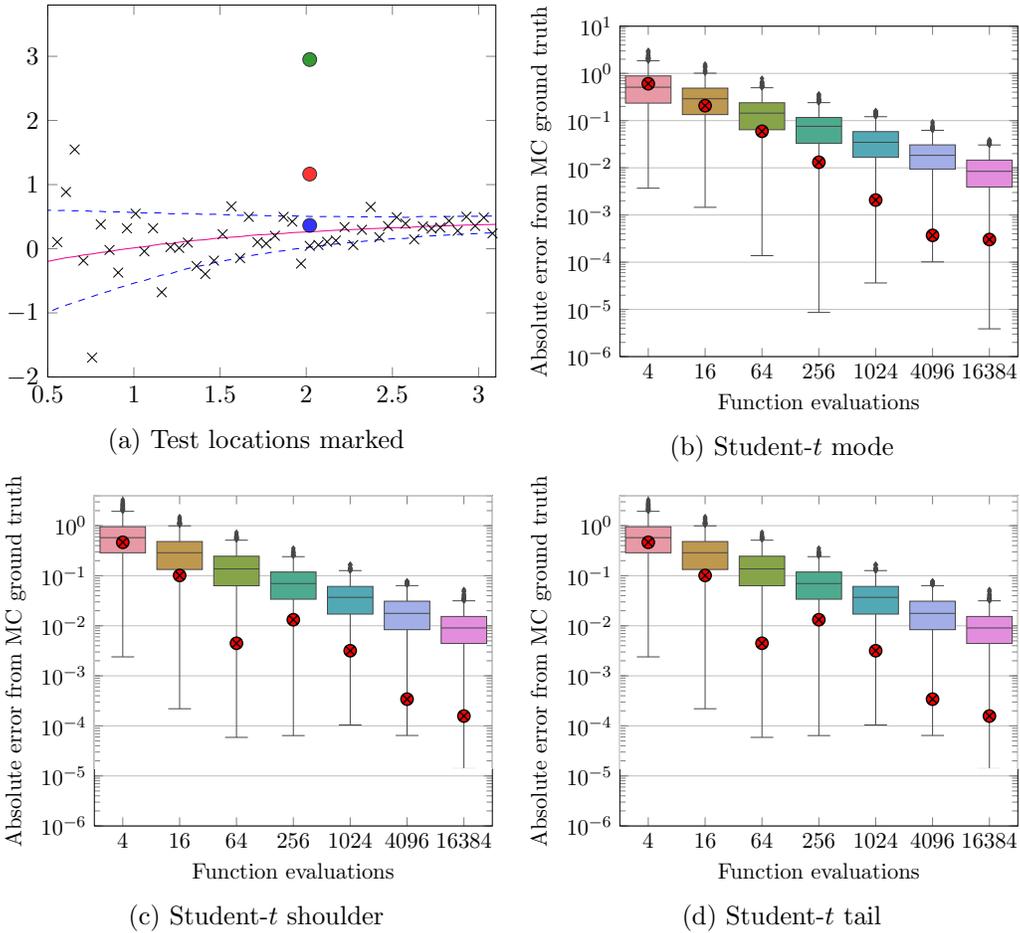

    \centering
    \begin{minipage}{.45\textwidth}
        \centering
        \resizebox{\textwidth}{!}{%
            \input{MCvsQuad_stut_process.tikz}
        }
        \subcaption{Test locations marked}
        \label{fig:stutprocess}
    \end{minipage}
    \begin{minipage}{.45\textwidth}
        \centering
        \resizebox{\textwidth}{!}{%
            \input{MCvsQuad_abslog_stut_mode.tikz}
        }
        \subcaption{Student-$t$ mode}
        \label{fig:stut_mode}
    \end{minipage}
    \begin{minipage}{.45\textwidth}
        \centering
        \resizebox{\textwidth}{!}{%
            \input{MCvsQuad_abslog_stut_shoulder.tikz}
        }
        \subcaption{Student-$t$ shoulder}
        \label{fig:stut_shoulder}
    \end{minipage}
    \begin{minipage}{.45\textwidth}
        \centering
        \resizebox{\textwidth}{!}{%
            \input{MCvsQuad_abslog_stut_far_tail.tikz}
        }
        \subcaption{Student-$t$ tail}
        \label{fig:stut_tail}
    \end{minipage}
    %\resizebox{0.45\textheight}{!}{
        %\input{MCvsQuad_abslog.tikz}
    %}
    \caption{Two dimensional Gauss-Hermite quadrature vs Monte Carlo. Each plot shows the log absolute error in estimating the two dimension integral required by our Heteroscedastic Student-$t$ model (see section~\ref{sec:nonGaussian}). In each case, the bias introduced by quadrature (circles) is small: a long way into the tail of the variance from the MC approximation. In fact, for small numbers of quadrature points, we often do better than the expected value using many more MC samples. Boxplots shows the absolute error on 1000 separate reruns of MC, whereas quadrature is deterministic. The error was evaluated at various points in the tail of the distribution as shown in \protect\subref{fig:stutprocess}).}
    \label{fig:MCvsQuad}
\end{figure}

Computing the expected likelihood requires many low-dimensional integrals. Recently, there has been progress in using stochastic methods to obtain unbiased estimates in this area using centered representations~\citep{Kingma:VB14, Rezende:stochastic14}. In this section, we re-examine the effectiveness of Gauss-Hermite quadrature in this setting. 
Gauss-Hermite quadrature approximates Gaussian integrals in one dimension using a pre-defined grid. For expectations of polynomial functions, the method is exact when the grid size meets the degree of the polynomial; for non-polynomial functions as we will encounter in general, we must accept a small amount of bias. To integrate higher dimensional functions, we must nest the quadrature, doing an integral across one dimension for each quadrature point in the other. Our experiments suggest that even in this case, the amount of bias is negligible, as Figure~\ref{fig:MCvsQuad} investigates, examining the accuracy of nested quadrature as compared to Monte Carlo estimates using the centered parameterization \citep{Kingma:VB14}. Inspired by an examination of quadrature for expectation propagation~\citep{Jylanki:stut11}, we examine the effectiveness for a several positions of the integral of a Student-$t$. 

Gauss-Hermite quadrature is appropriate for our integral as the Gaussian posteriors $q(\fV_{i})q(\gV_{i})$ are convolved with a function $p(\yV_{i}|\gV_{i},\fV_{i})$. Monte Carlo integration is exact in the limit of infinite samples, however in practice a subset of samples must be used. %Figure~\ref{fig:MCvsQuad} investigates the effectiveness and bias of the two dimensional quadrature used in the experiments.%, as a function of evaluations of $p(\yV|\fV,\gV)$. 
Gauss-Hermite requires ${\dataDim}h^\numLatentFuncs$ evaluations per point in the mini-batch, where $h$ is the number of Gauss-Hermite points used, $\dataDim$ is the number of output dimensions, and $\numLatentFuncs$ is the number of latent functions. 
%The figure shows the results of computing the $\int q(\fV_{i})q(\gV_{i})\log p(\yV_{i}|\fV_{i},\gV_{i})d\fV_{i}\,d\gV_{i}$ for a Student-$t$ likelihood~\eqref{eq:stut} (see \refsec{sec:nonGaussian}) and for a random input point $\xV_{i}$. 
Since Monte Carlo is unbiased, using a stochastic optimizer with the stochastic estimates of the integral and its gradients will work effectively~\citep{Nguyen:blackbox14,Kingma:VB14}, though we find the bias introduced by the quadrature approach to be negligible. For higher number of latent functions it may be more efficient to make use of low variance Monte Carlo estimates for the integrals.
Gradients for the model can be computed in a similar way with the Gaussian idenities used by~\citet{Opper:variational09}.

% In the next section we compare Monte Carlo and quadrature for the two dimensional case.
%\subsection{Signal variance function}
%Exploiting a re-parameterization it possible to learn a process for the noise variance, signal variance and mean function~\citep{Tolvanen:hetero14} with a Gaussian likelihood. Our model is also capable of using this re-parameterization to learn a sparse approximation to this. The ability to scale effectively with the data is important as the more latent functions we introduce, the more difficult they may be to determine, and thus the more data required. In this case the problematic integrals are analytically tractable.

\subsection{Gradients and Optimization}
\label{app:gradsandop}
Gradients can be computed similarly to~\citep{Hensman:class15} using the equalities,
\begin{align}
    \frac{\partial}{\partial\mu}\E{\gaussianDist{x}{\mu}{\sigma^{2}}}{f(x)} &= \E{\gaussianDist{x}{\mu}{\sigma^{2}}}{\frac{\partial}{\partial x}f(x)} \label{eq:derivmean}\\
    \frac{\partial}{\partial\sigma^{2}}\E{\gaussianDist{x}{\mu}{\sigma^{2}}}{f(x)} &= \frac{1}{2}\E{\gaussianDist{x}{\mu}{\sigma^{2}}}{\frac{\partial}{\partial x^{2}}f(x)} \label{eq:derivvar}
\end{align}
and the chain rule.

Since our posterior assumes factorization between $q(\fV)$ and $q(\gV)$ we simply do the gradients independently. That is calculate 
\begin{gather*}
    \frac{\partial}{\partial\muVf}\E{\gaussianDist{\xV_{i}}{\mVf}{\vVf}}{\log p(\yV|\fV,\gV)}\\
    \frac{\partial}{\partial\muVg}\E{\gaussianDist{\xV_{i}}{\mVg}{\vVg}}{\log p(\yV|\fV,\gV)}\\
    \frac{\partial}{\partial\vVf}\E{\gaussianDist{\xV_{i}}{\mVf}{\vVf}}{\log p(\yV|\fV,\gV)}\\
    \frac{\partial}{\partial\vVg}\E{\gaussianDist{\xV_{i}}{\mVg}{\vVg}}{\log p(\yV|\fV,\gV)},
\end{gather*}
independently using~\eqref{eq:derivmean} and~\eqref{eq:derivvar}. The expectations can then be done using quadrature, or Monte Carlo sampling.
As before 
\begin{align*}
    \mVf &= \Kfu\Kuuif\muVf\\
    \vVf &= \Kff + \Kfu\Kuuif(\Sf - \Kuuf)\Kuuif\Kuf\\
    \mVg &= \Kgu\Kuuig\muVg\\
    \vVg &= \Kgg + \Kgu\Kuuig(\Sg - \Kuug)\Kuuig\Kug.
\end{align*}
We then can chain using $\frac{\partial}{\partial\mVf}\E{\gaussianDist{\xV_{i}}{\mVf}{\vVf}}{\log p(\yV|\fV,\gV)}\frac{\partial\mVf}{\partial \Kfu}\frac{\partial \Kfu}{\partial \theta}$, where $\theta$ is a hyper parameter of the kernel $\Kf$. Similar chain rules can be written for the other derivatives.

The model contains variational parameters corresponding to $q(\uVf) = \gaussianDist{\uVf}{\muVf}{\Sf}$ and $q(\uVg) = \gaussianDist{\uVg}{\muVg}{\Sg}$ and the latent input locations, $\zM$. As such the parameters do not scale with $\numData$. Naively the number of parameters is $\mathcal{O}(\numLatentFuncs(\numInducing^{2} + \numInducing) + \numInducing)$ however we can reduce this to $\mathcal{O}(\numLatentFuncs(\frac{\numInducing^{2}}{2} + \numInducing))$ by parameterizing the Choleksy of the covariance matrices, $\Sf = L_{\fV}L_{\fV}^{\top}$ and $\Sg = L_{\gV}L^{\top}_{\gV}$. This has the added benefit of enforcing that $\Sf$ and $\Sg$ are symmetrical and positive definite.

We initialize the model with random or informed lengthscales within the right region, $\muVf$ and $\muVg$ are assigned small random values, $\Sg$ and $\Sf$ are given an identity form.
In practice during optimization we find it helpful to initially fix all the kernel hyperparameters and $\zM$ at their initial locations, optimize for a small number of steps, then allow the optimization to run freely. This allows the latent means $\muVf$ and $\muVg$ to move to sensible locations before the model is allowed to completely change the form of the function through the modification of the kernel hyperparameters. True convergence can be difficult to achieve due to the potentially number of strongly dependent parameters and the non-convex optimization problem, and in practice we find it helpful to monitor convergence. It is important to note however that the number of parameters to be optimized is \emph{fixed} with respect to $\numData$.

\subsection{Further Twitter experiment details}
The model used to model the twitter data has some interesting properties, such as the ability to model a transition from a unimodal distribution to a bimodal distribution.
The following plot shows how the distribution changes throughout time for the Labour dataset.

\begin{center}
    \includegraphics[width=.5\textwidth,keepaspectratio,trim={1.2in 0.5in 0.65in 1in},clip]{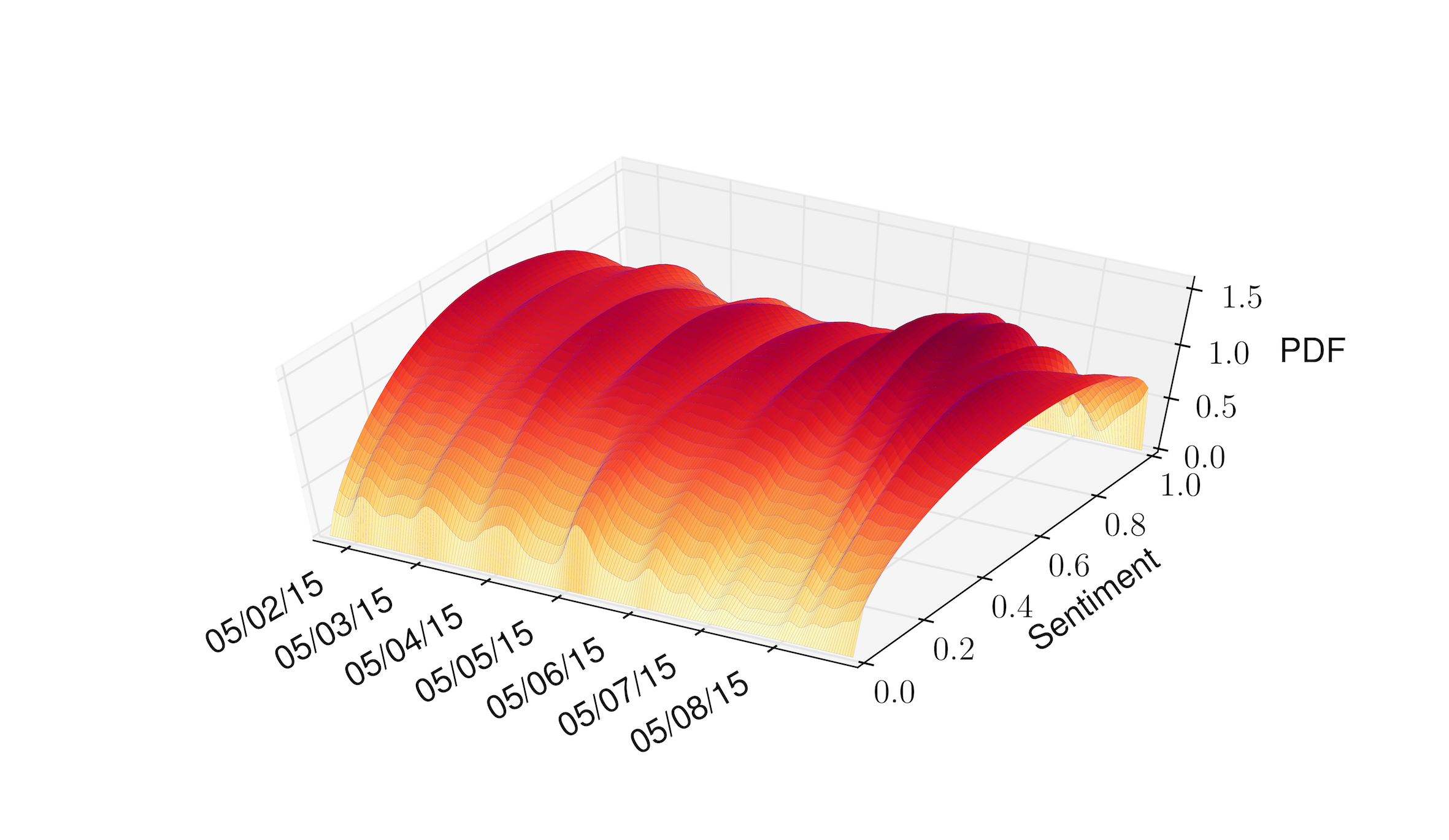}
\includegraphics[width=.5\textwidth,keepaspectratio]{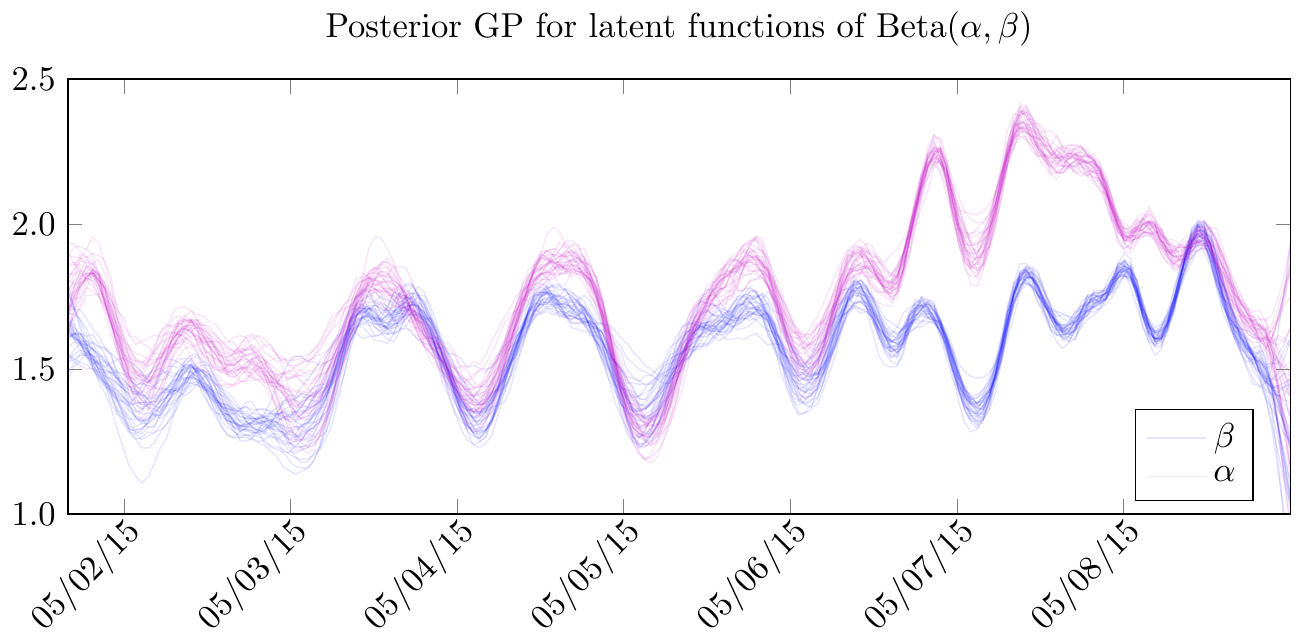}
    %\resizebox{0.5\textwidth}{!}{%
        %\input{labour_latent.tikz}
    %}
\end{center}

The latent functions $\alpha$ and $\beta$ which are modelled in Section~\ref{sec:twitter} can be plotted themselves. If both latent functions went below $1.0$ then the distribution at that time would turn into a bathtub shape. If both are larger than one but one is larger than the other, we have a skewed distribution. If one is below zero and the over above, it appears exponential or negative exponential.

\subsection{Survival details}
\label{appendix:survival}
To generate the synthetic survival dataset we first define latent functions that we wish to infer. These are a complex function of an input, $\xV$, with two dimensions,

\begin{align*}
    \alphaV &= \exp\left(2\exp(-30(\xV_{:,0} - \frac{1}{4})^2) + \sin(\pi\xV_{:,1}^2) - 2\right)\\
    \betaV &= \exp\left(\sin(2\pi\xV_{:,0}) + \cos(2\pi\xV_{:,1})\right).
\end{align*}

We then make 1000 synthetic individuals, with covariates sampled uniformly from $\xV_{i,0} \sim \text{Uniform}(0,1)$ and $\xV_{i,1} \sim \text{Uniform}(0,1)$.

Using these two latent functions, $\alphaV_{i}$ and $\betaV_{i}$, computed using covariates $\xV_{i}$ for individual $i$, we sample a simulated failure time from a log-logistic distribution,

\begin{equation*}
    \yV \sim LL(\alphaV, \betaV) = \frac{\left(\frac{\betaV}{\alpha}\right)\left(\frac{\yV}{\alpha}\right)^{\betaV - 1}}{\left(1 + \frac{\yV}{\alpha}^{\betaV}\right)^2}.
\end{equation*}

These are then the true failure times of individuals with covariates $\xV_{i}$. $20\%$ of the data is chosen to be censored censor. A time is uniformly drawn%, $\tTimeV_{i} \in [0, \yV_i]$
, and the observed time is truncated to this time, $\yV_{i} = \tTimeV_{i}$. Otherwise $\tTimeV_{i} = \yV_{i}$. Additionally a indicator $\censorV_{i} = 1$ is provided to the model if censoring occurs, and $\censorV_{i} = 0$ if the real failure time was observed. This mimics patients dropping out of a trial, with the assumption that the time at which they drop out is independent of the failure time and covariates. For these censored times, we only know that $\rTimeV_{i} > \tTimeV_{i}$, and for the uncensored individuals it is known that $\rTimeV_{i} = \tTimeV_{i}$. 

As such the likelihood is decomposed into $P(\tTime_{i} \leq \yV_{i} < \tTimeV_{i} + \delta\tTime|\alphaV_{i}, \betaV_{i}, \censorV_{i}=0)$ and $P(\yV_{i}|\alphaV_{i}, \betaV_{i}, \censorV_{i}=1) = 1 - P(\yV_{i}>\tTimeV_{i}|\alphaV_{i}, \betaV_{i}, \censorV_{i}=1)$

\begin{equation*}
    p(\yV|\alphaV, \betaV, \censorV) = \prod^{K : \censor \neq 1}_{i}\frac{\left(\frac{\betaV_{i}}{\alpha_{i}}\right)\left(\frac{\yV_{i}}{\alpha_{i}}\right)^{\betaV_{i} - 1}}{\left(1 + \frac{\yV_{i}}{\alpha_{i}}^{\betaV_{i}}\right)^2} \prod^{M : \censor = 1}_{j} \frac{1}{1 + \left(\frac{\yV_{j}}{\alpha_{j}}\right)^{\betaV_{j}}}
\end{equation*}

The task is then to infer $\alphaV$ and $\betaV$, such that we know how the failure time distribution varies in response to covariate information.

%Backup of full table
%\begin{table}[h]
%\label{table:dataset}
%\begin{tabular}{l l l l l||l l l l}
%\hline
 %\multicolumn{1}{c}{\textbf{Dataset}} & \multicolumn{4}{c||}{\textbf{MAE}} & \multicolumn{4}{c}{\textbf{NLPD}}\\ 
 %\multicolumn{1}{c}{} & \multicolumn{1}{c}{\textbf{G}} & \multicolumn{1}{c}{\textbf{Chained GP}} &\multicolumn{1}{c}{\textbf{StuT}} & \multicolumn{1}{c||}{\textbf{SVHStuT}} &
%\multicolumn{1}{c}{\textbf{G}} & \multicolumn{1}{c}{\textbf{Chained GP}} & \multicolumn{1}{c}{\textbf{StuT}} & \multicolumn{1}{c}{\textbf{SVHStuT}} \\
%\hline
%%Yuan200 & - & - & - & - & - & - & - & - \\ %http://www.machinelearning.org/proceedings/icml2007/papers/326.pdf talks about this dataset, essentially a sample from a heteroscedastic model
%Elevators1000 & 0.252 & 0.272 & - & - & 0.320 & 0.234 & - & - \\ %This is based on the three folds that worked for both models (initialization issues) at the minute, will change when experiments are rerun
%Elevators10000 & 0.219 & 0.224 & - & - & 0.193 & 0.111 & - & - \\ 
%%Friedman & - & - & - & - & - & - & - & - \\ 
%%FriedmanCluster & - & - & - & - & - & - & - & - \\  %This is the friedman dataset but with clustered outliers as in Twinned GPs
%CorruptMotor & 0.947 & 0.841 & 0.835 & 0.834 & 1.943 & 1.429 & 1.422 & 1.403 \\ %Corrupted silverman motorcycle dataset, with extra noisy elements added
%Boston & 0.048 & 0.050 & 0.050 & 0.050 & -1.207 & -1.341 & -1.261 & -1.352 \\
%\hline
%\end{tabular}
%\caption{Results showing the MAE and NLPD over 5 cross validation folds}
%\end{table}

\end{document}